\PassOptionsToPackage{table}{xcolor}
\documentclass{article}
\usepackage{iclr2027_conference,times}

\usepackage{amsmath,amsfonts,bm}

\def\eqref#1{equation~\ref{#1}}

\def\1{\bm{1}}

\DeclareMathAlphabet{\mathsfit}{\encodingdefault}{\sfdefault}{m}{sl}
\SetMathAlphabet{\mathsfit}{bold}{\encodingdefault}{\sfdefault}{bx}{n}

\usepackage{amsmath,amssymb}
\usepackage{graphicx}
\usepackage{booktabs}
\usepackage{array}
\usepackage{xcolor}
\usepackage{microtype}
\usepackage{hyperref}
\usepackage{url}
\iclrfinalcopy
\hypersetup{colorlinks=true,linkcolor=blue,citecolor=blue,urlcolor=blue}

\definecolor{circuitblue}{HTML}{4F86A5}
\definecolor{storeamber}{HTML}{B58A56}
\definecolor{useviolet}{HTML}{8B7BA7}
\definecolor{tableslate}{HTML}{7E8C99}
\definecolor{tableline}{HTML}{CBD4DD}
\definecolor{neutralwash}{HTML}{EDF1F4}
\definecolor{circuitwash}{HTML}{EAF2F8}
\definecolor{storewash}{HTML}{FBF1E2}
\definecolor{usewash}{HTML}{F1ECF6}
\definecolor{positivewash}{HTML}{EAF5EE}
\definecolor{negativewash}{HTML}{FAECEC}
\arrayrulecolor{tableline}
\newcommand{\neutralhead}{\rowcolor{neutralwash}}
\newcommand{\circuithead}{\rowcolor{circuitwash}}
\newcommand{\storehead}{\rowcolor{storewash}}
\newcommand{\usehead}{\rowcolor{usewash}}
\newcommand{\papertable}{%
  \normalsize
  \renewcommand{\arraystretch}{1.08}%
  \setlength{\tabcolsep}{5pt}%
}
\newcolumntype{N}[1]{>{\raggedleft\arraybackslash}p{#1}}
\newcolumntype{P}[1]{>{\raggedright\arraybackslash\hyphenpenalty=10000\exhyphenpenalty=10000}p{#1}}
\makeatletter
\newcommand{\paperneedspace}[1]{%
  \par\begingroup
  \@tempdima=#1\relax
  \vskip 0pt plus \@tempdima
  \penalty -100
  \vskip 0pt plus -\@tempdima
  \vskip \@tempdima
  \penalty 9999
  \vskip -\@tempdima
  \vskip 0pt
  \endgroup
}
\makeatother
\newcounter{paperalgorithm}
\newenvironment{paperalgorithm}[2]{%
  \refstepcounter{paperalgorithm}\label{#2}%
  \begin{center}\begin{minipage}{0.96\linewidth}
  \setlength{\fboxsep}{4pt}%
  \colorbox{neutralwash}{\parbox{\dimexpr\linewidth-2\fboxsep\relax}{%
    \textbf{Algorithm \thepaperalgorithm}\quad #1}}%
  \vspace{2pt}{\color{tableslate}\hrule}\vspace{4pt}\normalsize
  \begin{tabular}{@{}>{\color{tableslate}}r@{\hspace{9pt}}>{\raggedright\arraybackslash}p{\dimexpr\linewidth-16pt\relax}@{}}%
}{%
  \end{tabular}\vspace{4pt}{\color{tableslate}\hrule}
  \end{minipage}\end{center}%
}
\newcommand{\algline}[2]{#1 & #2\\[2.2pt]}
\newcommand{\circuit}{\textsc{Circuit}}
\newcommand{\store}{\textsc{Store}}
\newcommand{\usephase}{\textsc{Use}}

\newcommand{\jointb}{Joint-to-matching}
\newcommand{\ccur}{\textsc{CCUR}}
\newcommand{\mwrite}{\textsc{M-Write}}
\newcommand{\muse}{\textsc{M-Use}}

\title{What Should Data Teach? Moving Bottlenecks Across Circuit, Store, and Use}

\author{%
Yixiao Chen\textsuperscript{1,2,\dag}\quad
Ke Cheng\textsuperscript{1,*}\quad
Yue Liu\textsuperscript{1}\quad
Jun Zhang\textsuperscript{1}\quad
Jie Jiang\textsuperscript{1}\\[3pt]
\normalfont\textsuperscript{1}Tencent\quad
\textsuperscript{2}Harvard University\\[3pt]
\normalfont\texttt{\href{mailto:cyixiao1019@gmail.com}{cyixiao1019@gmail.com}}\quad
\texttt{\href{mailto:ckpassenger@buaa.edu.cn}{ckpassenger@buaa.edu.cn}}%
}
\hypersetup{%
  pdftitle={What Should Data Teach? Moving Bottlenecks Across Circuit, Store, and Use},
  pdfauthor={Yixiao Chen; Ke Cheng; Yue Liu; Jun Zhang; Jie Jiang}%
}

\begin{document}
\maketitle
\begingroup
\renewcommand{\thefootnote}{\fnsymbol{footnote}}
\footnotetext[1]{Corresponding author.}
\footnotetext[2]{Work done during an internship at Tencent.}
\endgroup
\setcounter{footnote}{0}

\begin{abstract}
What should data teach a language model at a particular point in training? A circuit view reveals three distinct bottlenecks: forming a computation, making its required content available, and selecting among available routes. A shared diagnosis-to-data principle connects them: localize the missing operation, preserve its causal relation, vary shortcut-bearing context, and re-audit the residual. Formation-sensitive selection and prerequisite ordering accelerate a binding--matching--transport path; a brief early prefix from the same training multiset retains a validation advantage through 100B tokens. Availability counterfactuals then distinguish writing content from invoking available memory, while paired supervision and context-opportunity ranking improve matched route decisions and long-context answer likelihood. A continuous 350M-model experiment connects the three interventions on the same facts: early circuit training improves subsequent learning, and the complete sequence outperforms stage-replacement controls on facts withheld from Use teaching. Independent query surfaces and opposed-source decisions expose conditional arbitration as the remaining frontier. Together, these results show why a change in the limiting operation calls for a change in supervision, not merely a new ranking of difficult examples.
\end{abstract}

\section{Introduction}

What limits a language model at a particular checkpoint? We began by asking whether the required computation had formed. Formation-time circuit signals organized an early prefix that accelerated a binding--matching--transport path. In the auxiliary order study, the same-multiset intervention improved Overall, Math, Code, and a natural induction-sensitive bucket, but not Web. On frozen Web cases, supplying the relevant evidence as context QA substantially improves target accuracy over source-masked and shuffled controls in both arms (Figure~\ref{fig:circuit-bridge}). The read path can use accessible evidence, raising the next diagnostic question: can model state supply the content the task requires?

This led from Circuit to Store. Availability counterfactuals separated targets that parametric memory could not supply from those available in isolation but suppressed in a natural carrier: writing and invoking memory were different operations. The latter exposed Use, where multiple information or computation routes are available and the query must determine which one controls the answer. Controlled supervision strengthens matched route decisions, but general arbitration across independently phrased and opposed routes remains difficult.

Our central claim is that the useful training target changes as a limiting operation is repaired: computation formation, content availability, and control among available routes call for different supervision. Circuit, Store, and Use are operational regimes rather than physical modules or a universal training clock. Natural-text residuals test what remains after formation; available-source conflicts test what remains once content can be supplied.

One diagnosis-to-data principle connects the chain: localize the missing operation, preserve its defining relation, vary shortcut-bearing context, and re-audit the remainder. We first isolate the demands and their repairs, then connect them in a continuous training experiment on the same facts. Stage-replacement controls test whether early circuit organization, target-fact writing, and paired Use supervision contribute to one final task (Section~\ref{sec:sequential}).

\begin{figure}[t]
  \centering
  \includegraphics[width=\textwidth]{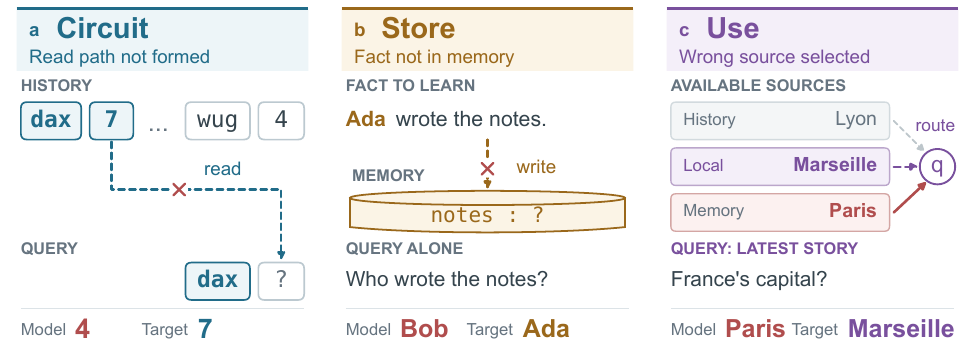}
  \caption{\textbf{Three failures, three teaching targets.} (a) Repeated \texttt{dax} should retrieve 7; the read fails. (b) ``Ada wrote the notes'' has not been acquired for unaided recall. (c) Query $q$ asks for the latest fictional story's capital, Marseille, but memory supplies Paris. Dashed paths denote candidate operations; crosses mark failed read/write steps; the solid red arrow marks wrong routing. Bottom entries compare model predictions (red) with target answers (panel colors). These are illustrative cases, not measured outputs.}
  \label{fig:overview}
\end{figure}

Figure~\ref{fig:overview} illustrates the three failures. H denotes remote history accessed through an induction-style read, L nearby context, and M parametric memory.

Our contributions are:
\begin{itemize}
  \item \textbf{A circuit-level reinterpretation of training bottlenecks.} Starting from Circuit formation, we show that residual failure can move from a missing computation to unavailable content and then to incorrect control among available routes. Circuit, Store, and Use are therefore operationally distinct failures rather than interchangeable notions of example difficulty.
  \item \textbf{A shared diagnosis-to-data principle with different repair actions.} Frozen failures, availability counterfactuals, and mechanism-sensitive signals identify the missing operation; relation preservation and context replacement then construct supervision suited to Circuit formation, Store acquisition or invocation, and matched Use decisions.
  \item \textbf{A discovery chain with sequential validation.} Early Circuit organization retains an advantage through 100B tokens. A separate continuous experiment connects Circuit, Store, and Use on common facts and improves final use over stage-replacement controls. Independent-query and opposed-source audits identify transferable conditional arbitration as the next bottleneck.
\end{itemize}

\section{Related Work}

\paragraph{Curriculum and adaptive selection.}
Curriculum and model-aware selection organize examples by difficulty, quality, prerequisites, or checkpoint-local influence \citep{chen2023skillit,wettig2024qurating,zhang2026beyond,zhang2025preference,yu2024mates,yu2025groupmates,smith2026influence,wang2025temporal,schoenegger2025influence}. Our distinction is not that data should change with training, but \emph{what supervision must change}: computation formation, content availability, and route selection require different repair actions rather than only a new ranking score.

\paragraph{Mechanistic data and circuit formation.}
MDA traces examples to interpretable units and accelerates induction-head convergence across Pythia scales \citep{chen2026mda}; Bi-Induct separates induction signatures from functional ICL gains \citep{sabry2026biinduct}. SMDA attributes training influence to symbolic behavioral policies \citep{habibi2026smda}, while SAMS schedules circuit-steered data by training utility \citep{lee2026sams}. Our distinction is the diagnosis of missing computation, unavailable content, and incorrect route control, and the different repair actions these require. MDA supplies the Circuit ranking signal; we test formation timing, prerequisites, exit, and persistence through 100B, then connect the repairs in a common training chain. Induction work motivates binding--matching--transport \citep{olsson2022induction,singh2024induction}; shortcut results motivate varied realizations \citep{kawata2025shortcut}.

\paragraph{Storage and incomplete learning.}
Knowledge exposure, extraction, and manipulation are distinct problems \citep{kandpal2023longtail,allenzhu2023storage,allenzhu2023manipulation,zhou2025cascade,xue2026incomplete}. Our Store category is functional: counterfactuals separate content that M-only cannot supply from content that M supplies but that loses control in a natural carrier, without asserting one physical memory location \citep{geva2021ffn}.

\paragraph{Long context and route control.}
Context--memory conflicts reveal task-dependent source reliance \citep{kortukov2024conflicts,xu2024conflicts,sun2026task}, while long-context selectors use attention or long--short loss contrasts to find informative windows \citep{chen2025ladm,wu2025longattn,fang2025longppl,deng2025longfilter}. Our clipped full--local response targets context opportunity; unlike confidence-weighted long--short selectors, its role here is a matched test of window identity. We distinguish that predictive dependence from learning which route a query authorizes, and separate matched-route and ranking gains from general arbitration across opposed routes.

\section{A Circuit View of Training}
\label{sec:diagnosis}

\subsection{From residual failure to missing operation}

Let $u=(x,y,q,S)$ denote context, target, query, and candidate sources. At checkpoint $\theta_t$, source-retain/remove/swap/truncation views assign a behavioral failure class $f_t(u)$. Separately, component sensitivity $G_c$ measures a gradient response, while $M_k$ aggregates the classified residual,
\begin{equation}
G_c(u,t)=\frac{\|\nabla_{\theta_c}\ell_t(u)\|_2^2}{|\theta_c|},
\qquad
M_k(t)=\sum_u\mathbf{1}[f_t(u)=k]\max\{\operatorname{NLL}_t(u)-2,0\},
\label{eq:diagnosis}
\end{equation}
where $M_k$ is excess-loss mass on fixed identities. Counterfactual gates determine $f_t$, not the magnitude of $G_c$. Gradients prioritize components within the gated population; controlled binding, matching, transport, ablation, and swap tests assess function. MDA subsequently ranks training examples for an identified Circuit role.

\circuit{} has an unambiguous source but no usable read path. \store{} lacks required M-only content. \usephase{} has available routes but the task-authorized one fails to control output. The Store study tests this boundary: once M-only succeeds, carrier suppression is a Use failure. Local denotes residual explained by nearby context in the natural-case audit; it is a descriptive category, distinct from arbitration between competing sources. Gates and precedence appear in Appendices~\ref{app:protocol} and~\ref{app:method-details}.

\subsection{From diagnosis to data}

Separate an operation schema $r$ from a context realization $z$:
\begin{equation}
u=(r,z),\qquad y=\pi_r(z),\qquad
\mathcal D_r^Q=\{(r,z_i,\pi_r(z_i)):z_i\sim Q(\cdot\mid r)\}_{i=1}^{n}.
\label{eq:construction}
\end{equation}
Construction holds schema $r$ fixed while replacing realization $z$: a role-complete read schema, a cue--value relation, or a query-authority rule. $P(\cdot\mid r)$ denotes task-valid contexts; $Q$ is the chosen training distribution with support within $P$. Each answer follows $\pi_r$.

Selection scores enter only after this operation gate. Formation-time MDA is one ranking signal for a missing Circuit role, whereas the full--local response ranks windows where remote context can change prediction. Their teaching meaning comes from the gated operation and relation-preserving controls, not score magnitude alone.

Concentration raises the density of $r$; variation prevents token, carrier, domain, or source identity from naming the answer without performing it. Table~\ref{tab:common-template} shows how the principle yields different actions and residuals.

\begin{table}[!h]
\caption{\textbf{The residual-bottleneck chain in our investigation.} Each intervention addresses an identified missing operation; the remaining bottleneck motivates the next stage.}
\label{tab:common-template}
\begin{center}
\fontsize{8}{8.8}\selectfont
\renewcommand{\arraystretch}{1.0}
\setlength{\tabcolsep}{2pt}
\begin{tabular}{@{}P{0.075\textwidth}P{0.17\textwidth}P{0.225\textwidth}P{0.23\textwidth}P{\dimexpr0.30\textwidth-16pt\relax}@{}}
\toprule
\neutralhead
Stage & Missing operation & Diagnostic & Intervention & Residual bottleneck $\rightarrow$ \\
\midrule
\rowcolor{circuitwash}\circuit{} & path not formed & component gradients / formation-time MDA & mechanism-matched ordering & content unavailable $\rightarrow$ \textbf{Store} \\
\rowcolor{storewash}\store{} & required content unavailable & M-only / evidence counterfactuals & Write; test invocation at the boundary & selection conflict $\rightarrow$ \textbf{Use} \\
\rowcolor{usewash}\usephase{} & which circuit/source wins & query-conditioned routing probes & paired routing / \ccur{} & multi-arm arbitration $\rightarrow$ \textbf{open} \\
\bottomrule
\end{tabular}
\end{center}
\end{table}

The primary 1.485B model runs through 100B tokens; auxiliary 160M studies isolate order. Store and Use studies start from the completed 100B model. A separate 350M chain applies the three interventions sequentially to a common factual task (Section~\ref{sec:sequential}). Appendix~\ref{app:setup} identifies each model and comparison.

\subsection{Decision-changing variation}

For a feasible context block $j$ (grouping variables that must change together), define
\begin{equation}
j\in\mathcal J_r
\iff \exists z,z'\in\operatorname{supp}P(\cdot\mid r):
z_{-j}=z'_{-j}\ \land\ \pi_r(z)\neq\pi_r(z').
\label{eq:decision-changing}
\end{equation}
Here $z_{-j}$ fixes the remaining blocks. In an opposed-source pair, hold the fact, passage, and candidate values fixed; change the task-authority instruction as one block. If this switches the required answer, that block belongs to $\mathcal J_r$. Equivalent paraphrases should instead preserve the answer. Equation~\ref{eq:decision-changing} specifies feasible construction contrasts, not an estimator of a latent language distribution.

Circuit varies tokens and domains around a role graph; Store varies cues around an assignment and audits wrong-pair training. Use must cover opposed authority decisions as well as equivalent wordings: rewarding one source throughout permits a marginal preference instead of the conditional rule.

\subsection{How the bottleneck moved in our investigation}

\begin{figure}[t]
  \centering
  \includegraphics[width=\textwidth]{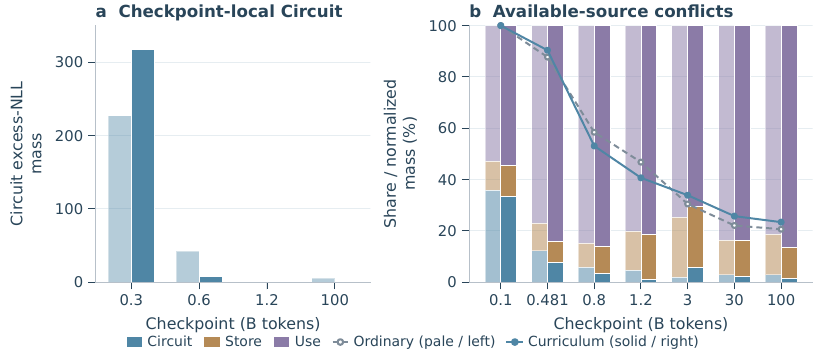}
  \caption{\textbf{Circuit residual contracts; available-source conflicts persist.} (a) Circuit excess-NLL mass from checkpoint-local diagnosis on 272 fixed natural identities. (b) Composition of a separate 942-target conflict residual; lines normalize total mass to each arm's 0.1B value. Pale/left bars: Ordinary; solid/right: Curriculum. Appendix~\ref{app:protocol} gives roster construction, rules, and sensitivity.}
  \label{fig:diagnosis}
\end{figure}

On 272 fixed natural identities, checkpoint-local diagnosis shows earlier Circuit contraction under Curriculum (Figure~\ref{fig:diagnosis}a). A separate available-source conflict panel retains a Use-heavy residual after formation (b). These complementary audits motivate the next repair; Section~\ref{sec:sequential} then tests the interventions on one common training chain and factual panel.

\section{Circuit: Forming the Computation}
\label{sec:circuit}

\paragraph{Localization.}
Information is present, but the required computation has not formed. Controlled probes separate previous-token binding, content matching, and value transport. Binding becomes usable first, while top-ranked head identities change during formation; the invariant is the role-complete relation, not a head coordinate. Formation-time signals rank natural examples. Natural-text ambiguous recall---a repeated predecessor--target pair whose predecessor has multiple observed continuations in the available document prefix---tests transfer beyond the synthetic construct.

\paragraph{Formation intervention.}
\textbf{Fixed:} the role-complete binding--matching--transport schema. \textbf{Varied:} natural domains, sources, token identities, and lexical realizations. \textbf{Audit:} functional formation, teacher time, order, exit, and ordinary continuation. Let $\mathcal C_{\mathrm{match}}$ be the Q/K component set frozen by the held-out matching gate and $\mathcal D_d$ the candidate pool in domain $d$. The formation teacher is the within-domain selection
\begin{equation}
\mathcal S_{\mathrm{match}}(\tau,d)=\operatorname{TopK}_{z\in\mathcal D_d}
\operatorname{MDA}_{\mathcal C_{\mathrm{match}}}(z;\theta_\tau),
\label{eq:mda-teacher}
\end{equation}
where $\tau$ is the formation or endpoint checkpoint and the slot budget is fixed. MDA supplies the component-sensitive score; our intervention adds role decomposition, prerequisite order, and formation-guided exit. \jointb{} exposes the relation before concentrating matching demand (Appendix Figure~\ref{fig:circuit-construction}). A pilot identifies the formation window and selects the exit by held-out performance; this schedule is frozen before the 100B run.

\begin{figure}[t]
  \centering
  \includegraphics[width=\textwidth]{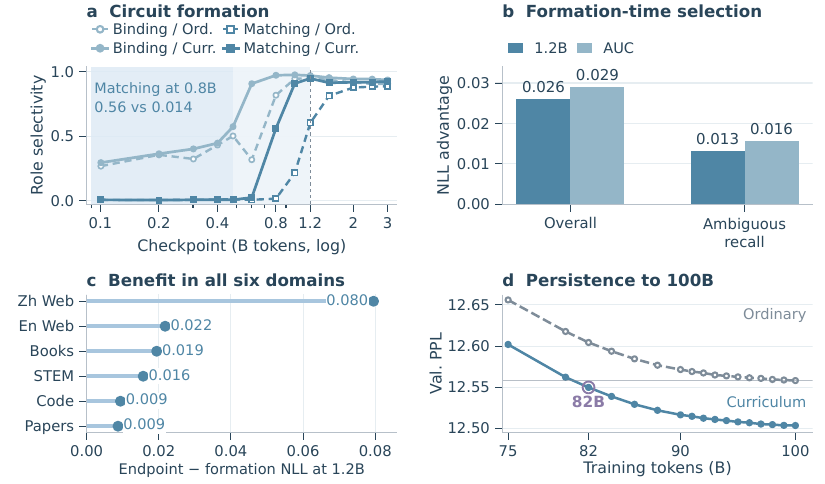}
  \caption{\textbf{Circuit training advances formation with persistent gains.} (a) Binding forms in both arms; reordering advances matching. (b) Endpoint-minus-formation NLL is positive for Overall and ambiguous recall at 1.2B and over normalized 0.1--1.2B AUC; 1.2B PPL reductions are 2.57\% and 1.30\%. (c) The same strict matching-only teacher comparison favors formation-time selection in all six domains at 1.2B. (d) Curriculum crosses Ordinary's final PPL at 82B and stays lower through 100B. Appendix~\ref{app:circuit} gives teacher trajectories, order controls, and the measured-exit sweep.}
  \label{fig:circuit}
\end{figure}

\begin{figure}[t]
  \centering
  \includegraphics[width=\textwidth]{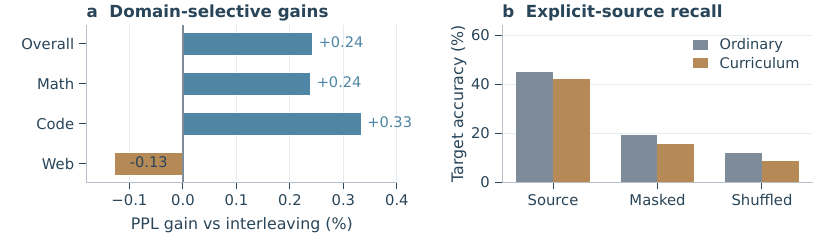}
  \caption{\textbf{Domain-selective gains motivate an availability audit.} (a) At 2.5B in the 160M order study, fixed-multiset \jointb{} improves Overall, Math, and Code PPL over interleaving, but not Web. (b) Relevant-source, masked-source, and shuffled-context QA on 435 fixed Web facts at 2.5B shows both arms use explicit evidence. Appendix Tables~\ref{tab:web-qa} and~\ref{tab:order-control} give full presentation/NLL comparisons and reverse-order controls, respectively.}
  \label{fig:circuit-bridge}
\end{figure}

\paragraph{Formation evidence.}
Curriculum advances matching while Ordinary still shows little matching selectivity, although binding is usable in both arms (Figure~\ref{fig:circuit}a). With matching-only organization fixed, formation-time MDA improves Overall and ambiguous recall prediction, with lower endpoint NLL in every domain (Figure~\ref{fig:circuit}b,c). This isolates teacher timing: the endpoint identifies a functioning path, whereas formation-time examples accelerate its construction. With membership fixed, \jointb{} beats interleaving and reverse order (Figure~\ref{fig:circuit-bridge}a; Appendix Table~\ref{tab:order-control}); the exit sweep peaks after the measured path forms (Table~\ref{tab:exit-sweep}). These controls separate organization from selection and calibrate exit around functional formation.

\paragraph{Persistent pretraining gains.}
Against matched interleaving, the same-multiset \jointb{} order improves Overall, Math, and Code PPL but not Web (Figure~\ref{fig:circuit-bridge}a; Appendix Table~\ref{tab:order-control}). Separately, Curriculum improves Overall PPL over Ordinary across three independent 3B runs, and both auxiliary runs pass the frozen binding, matching, and transport gates (Tables~\ref{tab:3b-replication} and~\ref{tab:abc-100m-confirm}). In the target-scale trajectory, Curriculum surpasses Ordinary's final validation performance at 82B rather than 100B tokens, then retains its advantage through the endpoint (Figure~\ref{fig:circuit}d). Every endpoint domain favors Curriculum (Table~\ref{tab:100b-domains}). The intervention uses a selected 1.2B-token prefix within the same total training multiset, followed by the ordinary sampling schedule.

MDA establishes that mechanistic augmentation accelerates circuit convergence \citep{chen2026mda}. We extend this to long-horizon general pretraining: to our knowledge, this is the first evidence that a circuit-sensitive early prefix retains a validation advantage after the intervention ends and ordinary training continues to 100B tokens.

\paragraph{Residual bottleneck.}
Despite the Web exception above, both arms answer the same frozen Web facts substantially better with relevant context than with a masked or shuffled source (Figure~\ref{fig:circuit-bridge}b). The read path is usable, motivating the next test: can parametric memory supply the required content? The separate 350M chain further shows that early Circuit training improves subsequent memory acquisition (Appendix~\ref{app:sequential}, Table~\ref{tab:sequential-deltas}), connecting read-path formation to the next intervention.

\section{Store: Making Content Available}
\label{sec:store}

\paragraph{Localization.}
The Web residual left by Circuit training raises a different question: given a usable computation, is the content it needs available in model state? We use Store as a functional availability category, not as a claim that one architectural module physically contains memory. Knowledge exposure, availability, and invocation are not equivalent. \mwrite{} targets cases where M-only cannot supply the value; \muse{} targets cases where M-only succeeds but the value loses control under a natural carrier. Let $m_\theta(u;x)$ be the correct-minus-competing answer log-probability margin at the first divergent answer token. For explicit-evidence, M-only, and natural-carrier views, define evidence rescue and circuit drag,
\begin{equation}
\Delta_E(u)=m_\theta(u;x_E)-m_\theta(u;x_M),
\qquad
D_{\mathrm{drag}}(u)=m_\theta(u;x_M)-\overline m_\theta(u;x_F).
\label{eq:memory-diagnosis}
\end{equation}
Here $\overline m_\theta$ averages the two natural carriers. The natural availability gate requires M-only candidate failure, evidence-present success, and sufficient evidence rescue. An invocation case instead requires M-only success and suppression under both answer-free carriers. These views separate missing content from suppressed content and generic difficulty (Appendix Table~\ref{tab:m-missing-gate}; protocols in Appendix~\ref{app:store-views}).

\paragraph{Repair intervention.}
\textbf{Fixed:} cue--value relation and answer. \textbf{Varied:} cue wording, carriers, and optimizer positions. \textbf{Audit:} relation direction, unseen cues, and natural invocation. \mwrite{} distributes independently supervised cue--value segments across updates. Wrong-pair training preserves answer-token exposure but changes the assignment. When M-only succeeds, \muse{} places the value behind the suppressing natural carrier. The same fact therefore calls for different supervision depending on availability (Appendix Figure~\ref{fig:store-construction}).

\paragraph{Relation acquisition.}
On controlled novel relations, the first test asks whether \mwrite{} learns the assignment rather than an answer-token prior. A low writing dose shifts the answer margin toward the intended value under both exact and unseen cues; wrong-pair training shifts it toward the swapped assignment (Appendix Tables~\ref{tab:m-write-dose} and~\ref{tab:relation-controls}). This assignment-sensitive response across cue forms demonstrates directional relation learning. We next test which supervision helps natural facts that are already available.

\paragraph{Acquisition versus invocation.}
The second test freezes 512 available but circuit-dragged facts and crosses two intervention slots; disabling an operation substitutes matched control facts rather than removing tokens. Use contributes more natural-carrier improvement than Write, with only a small interaction (Table~\ref{tab:store-main}; Appendix Table~\ref{tab:store-contrasts}). This difference also appears in candidate accuracy under alternate prompt B, which Use training does not see. Write raises M-only gain at either Use setting. Thus the operative distinction is what supervision helps available content control a prediction, not simply whether the model has encountered the fact.

\begin{table}[!h]
\caption{\textbf{Available facts benefit most from invocation training.} Parent-minus-continuation NLL gains ($\uparrow$); B accuracy is first-divergent-token candidate choice, not free generation. All arms share 512 facts, two matched slots, parent, and 25M budget. Use trains carrier A; natural gain averages A/B on those facts. Appendix~\ref{app:store} gives carrier and factorial contrasts.}
\label{tab:store-main}
\begin{center}
\normalsize
\setlength{\tabcolsep}{5pt}
\begin{tabular}{llrrr}
\toprule
\storehead
Arm & Active operation & M-only gain & Natural-use gain & B accuracy (\%) \\
\midrule
C0 & matched controls & 0.636 & 0.894 & 82.03 \\
C1 & Write & 0.740 & 0.971 & 83.98 \\
C2 & Use & 0.723 & 1.201 & 93.36 \\
C3 & Write + Use & \textbf{0.806} & \textbf{1.294} & \textbf{95.31} \\
\bottomrule
\end{tabular}
\end{center}
\end{table}

\paragraph{Residual bottleneck.}
In the common-fact chain, Store lowers memory NLL, and replacing target writing with control facts weakens final joint use (Appendix~\ref{app:sequential}, Tables~\ref{tab:sequential-stages} and~\ref{tab:sequential-full}). Store separates values that must be acquired from those that are available but fail to control prediction. The latter is no longer missing content; it is control over available information and computation routes. We therefore turn to Use: which route should the task authorize?

\section{Use: Selecting Among Available Routes}
\label{sec:use}

\paragraph{Localization.}
Once computation and content are available, Use asks which route should control the answer. History (H), Local context (L), and parametric Memory (M) may offer competing values. The residual in Figure~\ref{fig:diagnosis}b motivates teaching selection under the query and authority relation.

\paragraph{Application intervention.}
\textbf{Fixed:} the task-authorized winner rule or full--local opportunity criterion. \textbf{Varied:} query surface, facts, documents, sources, and candidates. \textbf{Audit:} independent surfaces, disjoint documents, and opposed routes. The invariant is not ``prefer H/L/M'': changing authority must change the winner when the sources disagree (Appendix Figure~\ref{fig:use-construction}).

\begin{figure}[t]
  \centering
  \includegraphics[width=\textwidth]{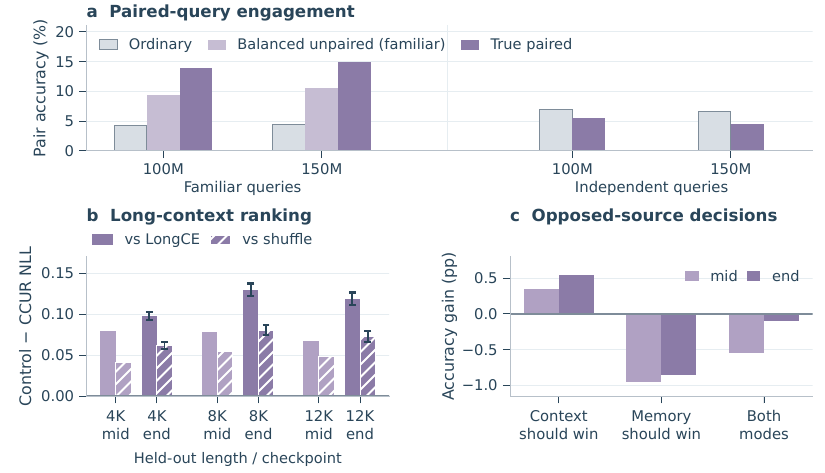}
  \caption{\textbf{Matched supervision improves engagement and long-context ranking.} (a) Paired supervision outperforms Ordinary and balanced unpaired on familiar queries; independent phrasings test transfer. (b) On a document-disjoint training pool, \ccur{} beats LongCE and score-shuffle; whiskers are paired 95\% endpoint intervals. (c) \ccur{} minus LongCE accuracy on opposed queries shows a source-prior shift, not joint arbitration. Positive values in (b,c) favor \ccur{}.}
  \label{fig:use}
\end{figure}

\paragraph{U1: matched-route engagement.}
True pairing teaches both authority modes per fact; balanced unpaired teaching assigns one mode per fact, balanced across facts. Both use ordinary language modeling plus the same auxiliary target CE and competitor-margin loss. True pairs share one per-fact template; unpaired decisions use the other two templates, so the comparison tests the complete constructions (Appendix Table~\ref{tab:paired-construction}). On unseen fact families with familiar queries, paired supervision improves strict-pair accuracy by 4.4 percentage points over balanced unpaired teaching at +100M (Figure~\ref{fig:use}a). On Store-taught facts withheld from Use teaching, the continuous chain likewise improves joint use after paired supervision and outperforms the unpaired replacement (Figure~\ref{fig:sequential}; Table~\ref{tab:sequential-main}; Appendix~\ref{app:sequential}). Reversed supervision reverses an independent route offset (Table~\ref{tab:use-diagnostics}), supporting relation engagement under matched query surfaces.

\paragraph{U2: task-aligned long-window ranking.}
For scored positions $I_x$ in a 16K window, Counterfactual Context-Use Ranking (\ccur{}) measures the clipped full--local response,
\begin{equation}
s_{\mathrm{CCUR}}(x)=\frac{1}{|I_x|}\sum_{i\in I_x}
\operatorname{clip}(\min\{\ell_i^L,\ell_i^R\}-\ell_i^F,0,\log 5),
\label{eq:ccur}
\end{equation}
where $\ell_i^F$ is full-window NLL; $\ell_i^L,\ell_i^R$ use the same local tokens with reset/original positions (Appendix~\ref{app:use}). \ccur{} and matched LongCE~\citep{fang2025longppl} share objective, source partition, quotas, updates, budget, and difficulty strata; only window identity changes within each cell. A score shuffle preserves the within-stratum score multiset but breaks its association with window identity. On a document-disjoint pool, \ccur{} lowers RULER-style answer-token NLL relative to both controls at every evaluated length and checkpoint, with paired endpoint intervals excluding zero throughout (Figure~\ref{fig:use}b; Appendix Table~\ref{tab:ccur-ci}).

\paragraph{Conditional arbitration as the next bottleneck.}
Matched relation engagement does not reliably transfer to independently phrased queries: expanding the training families still leaves strict-pair accuracy below Ordinary at both saved checkpoints (Appendix Table~\ref{tab:paired-failures}). \ccur{} likewise trades improved context-authorized decisions for worse memory-authorized decisions, without improving joint success (Figure~\ref{fig:use}c; Table~\ref{tab:hm-boundary}). These audits distinguish strengthening a route from learning when it should win: the remaining target is a query-conditioned decision that transfers across wording and reverses with authority.

Section~\ref{sec:sequential} assembles the stages into one continuous training chain; Appendix~\ref{app:use} gives the full transfer audits.

\section{Connecting the Stages}
\label{sec:sequential}

Do the three repairs contribute to one final task? Two 350M routes share initialization, the 3B pretraining multiset, and domain slots; only the early order differs. Each continues through 50M Store and 100M Use, inheriting its parent state. We track the same 640 facts, taught during Store but excluded from Use teaching, under fixed reading, memory, and opposed-authority queries. Appendix~\ref{app:sequential} gives the branches and complete outcomes.

\begin{figure}[!htbp]
  \centering
  \includegraphics[width=\textwidth]{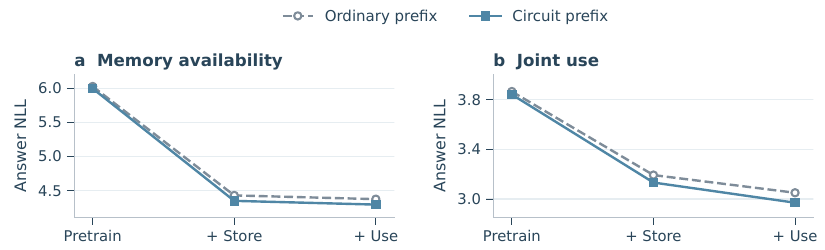}
  \caption{\textbf{The repairs compose on common facts.} Both routes receive target Store and paired Use after pretraining. On 640 Store-taught, Use-untaught facts, memory improves mainly during Store; the joint-use task improves further during Use. Each point evaluates the same fixed queries; lower answer-token NLL is better.}
  \label{fig:sequential}
\end{figure}

The early-course advantage extends into subsequent learning (Figure~\ref{fig:sequential}): Store reduces memory NLL more under the curriculum route, and Use produces a larger additional joint-task gain (Appendix Table~\ref{tab:sequential-deltas}). At endpoint, the complete sequence outperforms all four controls on joint-use NLL and strict-pair candidate accuracy (Table~\ref{tab:sequential-main}). Replacing target writing or paired teaching weakens final use despite retaining the other stages, connecting all three interventions through one common outcome.

\begin{table}[!htbp]
\caption{\textbf{Each stage contributes to final use.} Same 640 facts and 1,920 familiar-query pairs. Pair accuracy requires both authorized candidates to win at their first divergent token. ``Control facts'' replaces target Store; ``unpaired'' replaces the complete Use construction.}
\label{tab:sequential-main}
\begin{center}\normalsize
\setlength{\tabcolsep}{7pt}
\begin{tabular}{lrr}
\toprule\neutralhead
Training route & Joint-use NLL $\downarrow$ & Pair accuracy (\%) $\uparrow$ \\
\midrule
Complete sequence & \textbf{2.970} & \textbf{5.26} \\
No early curriculum & 3.050 & 3.39 \\
Control facts in Store & 3.245 & 1.41 \\
Unpaired Use & 3.019 & 3.18 \\
Ordinary continuation & 3.870 & 0.99 \\
\bottomrule
\end{tabular}
\end{center}
\end{table}

The worked case in Appendix Table~\ref{tab:sequential-case} makes the handoff concrete: the model can read Newton's death year before recalling it unaided; Store supplies recall, and Use subsequently resolves the illustrated opposed-query pair. The aggregate gain is on familiar instructions; the cross-surface arbitration frontier identified above remains.

\paperneedspace{8\baselineskip}
\section{Conclusion}

Circuit formation, content availability, and route control require different supervision. Our diagnostic studies identify these demands; a continuous common-fact experiment shows how their interventions compose, while the early Circuit advantage persists through 100B tokens in the long-horizon study. The remaining challenge is conditional arbitration that transfers across wording and opposed winners. The same principle guides that next step: preserve the required operation while varying the context that could substitute a shortcut.

\bibliography{references}
\bibliographystyle{iclr2027_conference}

\appendix
\addtocontents{toc}{\protect\appendixcontentsstart}
\clearpage
\section*{Appendix Contents}
\pdfbookmark[0]{Appendix Contents}{appendix.contents}
\begingroup
\setcounter{tocdepth}{2}
\hypersetup{linktoc=all}
\normalsize
\setlength{\parskip}{0pt}
\makeatletter
\let\l@section\@gobbletwo
\let\l@subsection\@gobbletwo
\newcommand{\appendixcontentsstart}{%
  \renewcommand{\l@section}[2]{%
    \addvspace{0.6em}%
    \@dottedtocline{1}{0em}{2.5em}{\bfseries ##1}{\bfseries ##2}%
  }%
  \renewcommand{\l@subsection}{\@dottedtocline{2}{1.25em}{3em}}%
}
\@starttoc{toc}
\makeatother
\endgroup
\clearpage

\makeatletter
\renewenvironment{table}[1][]{%
  \par\addvspace{\intextsep}\noindent
  \begin{minipage}{\linewidth}\def\@captype{table}%
}{%
  \end{minipage}\par\addvspace{\intextsep}%
}
\makeatother
\raggedbottom

\section{Shared Setup and Scope}
\label{app:setup}

\subsection{Primary scale run}

\begin{table}[!htbp]
\caption{Configuration of the matched 1.485B-parameter/100B-token experiment.}
\label{tab:config}
\begin{center}
\papertable
\setlength{\tabcolsep}{4.5pt}
\begin{tabular}{>{\bfseries}P{0.10\textwidth}P{0.18\textwidth}P{0.48\textwidth}P{0.13\textwidth}}
\toprule
\neutralhead
Block & Item & Specification & Pair status \\
\midrule
Model & Architecture & GPT decoder; 28 layers; $d=2{,}048$; SwiGLU 5,120 & matched \\
Model & Parameters & 1,484,900,352 & matched \\
Model & Attention & 16-head MHA; RoPE; SDPA & matched \\
Model & Norm / vocab & RMSNorm; 65,536 tokens & matched \\
Input & Sequence / batch & 2,048 tokens; 512 sequences (1,048,576 targets/update) & matched \\
Train & Optimizer & AdamW; $\beta=(0.9,0.95)$; weight decay 0.1 & matched \\
Train & Learning rate & $2\times10^{-4}$ peak; $2\times10^{-5}$ minimum & matched \\
Train & Schedule & approximately 300M warm-up; cosine decay to 100B & matched \\
Data & Fixed & initialization, multiset, domain slots, optimizer, LR, budget & matched \\
\rowcolor{circuitwash}Data & Changed & positions of the same examples & \textbf{only change} \\
\bottomrule
\end{tabular}
\end{center}
\end{table}

The Curriculum prefix places joint binding--matching examples before matching-selected examples; Ordinary encounters the same sequences elsewhere in the fixed multiset. Domain weights and validation targets are identical across arms. Table~\ref{tab:circuit-settings} records the prefix allocation and data sizes. The long-horizon experiment contains one training trajectory per arm; the early comparison uses three independent initializations (Table~\ref{tab:3b-replication}).

\begin{table}[!htbp]
\caption{Evidence map. Separate rows use separate populations; shared terminology does not merge them.}
\label{tab:model-map}
\begin{center}
\papertable
\setlength{\tabcolsep}{2.5pt}
\begin{tabular}{P{0.18\textwidth}P{0.19\textwidth}P{0.34\textwidth}P{0.18\textwidth}}
\toprule
\neutralhead
Evidence & Model / start & Matched object & Support \\
\midrule
Moving residual & 1.485B pair & 272 dynamic identities; 32 fixed-stratum cases; 942 conflicts & separate frozen panels \\
MDA teacher time & 350M proxy; 1.485B target & formation-time vs endpoint-time MDA; matching-only selection & 1 run per arm \\
Circuit confirmation & Pythia-style 160M @100M tokens & clean/swap, binding ablation, forced transport & 2 runs $\times$ 3 domains \\
Order & Pythia-style 160M & same selected multiset; 2.5B endpoint & 1 run per arm \\
Exit & 1.485B target & same multiset; 3B endpoint & 1 run per arm \\
Long horizon & 1.485B target & same 100B multiset, positions only & dense 75--100B trajectory \\
M availability & Ordinary@100B & 11,997 natural candidates & 2,220 M-missing \\
Memory action & Ordinary@100B & relation dose plus natural Write$\times$Use & 25M continuations \\
Paired Use & Ordinary@100B & familiar and independent query surfaces & matched continuations \\
Context ranking & Ordinary@100B & original and document-disjoint pools & matched continuations \\
Sequential validation & 350M; common initialization & 640 common facts; continuous Circuit--Store--Use & 5 final routes \\
\bottomrule
\end{tabular}
\end{center}
\end{table}

The Pythia-style suite \citep{biderman2023pythia} complements the target-scale run by isolating functional order under inexpensive same-multiset controls. The selector teacher is instead a custom 350M decoder sharing the primary model's tokenizer and packed candidate store (Table~\ref{tab:aux-config}). Transfer consists of example rankings and sequence identities, not model weights or Pythia token IDs. Store and Use continuations start from completed checkpoints; the separate sequential study is specified in Appendix~\ref{app:sequential}.

\begin{table}
\caption{Auxiliary architectures and the selector proxy. The Pythia suite and the primary pretraining pipeline have separate tokenizations.}
\label{tab:aux-config}
\begin{center}\papertable
\begin{tabular}{@{}P{0.26\textwidth}P{0.70\textwidth}@{}}
\toprule\neutralhead
Model & Configuration \\
\midrule
Pythia 160M auxiliary & GPT-NeoX; 12 layers; width 768; 12 heads; FFN 3,072; GELU; LayerNorm; official 50,304-token vocabulary and tokenizer \\
350M selector teacher & 349,746,176 parameters; 22 layers; width 1,024; 16-head MHA; SwiGLU 2,816; RMSNorm; RoPE; tied embeddings; primary 65,536-token tokenizer \\
Teacher training & 2,048-token sequences; effective batch 512; AdamW, $\beta=(0.9,0.95)$, weight decay 0.1; LR $3\times10^{-4}$ to $3\times10^{-5}$; 95 warm-up updates; cosine schedule to 3B \\
\bottomrule
\end{tabular}
\end{center}
\end{table}

\section{Common Diagnosis}
\label{app:protocol}

\subsection{Operational rule}

\begin{paperalgorithm}{Task-conditioned frozen-failure audit}{alg:failure}
\algline{1}{\textbf{Input:} a frozen target, query, candidate sources, checkpoint, and source-retain/remove/value-swap/local-truncation views.}
\algline{2}{Assign Circuit when a single unambiguous source exists but binding--matching--transport is unusable.}
\algline{3}{Assign Store when no reliable contextual source exists, local context is insufficient, and M-only does not supply the target.}
\algline{4}{Assign Use when source content is available but the query-required source or value fails to control the output.}
\algline{5}{Preserve Local, mixed, nonactionable, and ambiguous/invalid states; aggregate only actionable excess-loss mass.}
\algline{6}{Reapply the same rule at every checkpoint and report frozen-population scope, transitions, and cutoff sensitivity.}
\end{paperalgorithm}

For the conflict panel, let $p_v$ be target log probability in view $v$. Availability is $a=p_{\mathrm{correct\ only}}-p_{\mathrm{no\ source}}$; correct-source effect is $c=p_{\mathrm{full}}-p_{\mathrm{mask\ correct}}$; wrong-source removal gain is $w=\max_j(p_{\mathrm{mask\ wrong}_j}-p_{\mathrm{full}})$; and local gain is $l=\max(p_{\mathrm{local32}},p_{\mathrm{local128}})-p_{\mathrm{full}}$. Availability requires source-only success or a sufficient source-only gain. Actionability requires a remaining full-view failure and causal evidence from history or local context. Ordered rules then identify selection, local--global arbitration, value-use, and binding; unresolved ties remain ambiguous. Table~\ref{tab:classifier-rules} gives the exact gates and precedence.

Rescaling the positive strong-rule thresholds changes few assignments and leaves the actionable class composition nearly unchanged. The moving-residual result is stable across the registered range (Table~\ref{tab:audit-sensitivity}).

\subsection{Fixed-case trajectories}

The 32-case panel contains eight cases per retrospective, intervention-refined class. Its labels stay fixed while losses change: Table~\ref{tab:unified-longitudinal} measures learning within fixed strata, not checkpoint-local reclassification. Local cases retain or improve target support under local32/local128 truncation; Store cases lack a reliable source and are not explained by that local support. Mixed cases remain separate in the operational audit.

\begin{table}[!htbp]
\caption{Fixed retrospective strata on the same 32 natural cases. C/S/U/L are Circuit/Store/Use/Local excess-mass shares (\%); Mass is absolute excess NLL. Each group contains eight fixed identities and retains its label.}
\label{tab:unified-longitudinal}
\begin{center}
\papertable
\begin{tabular}{llrrrrr}
\toprule\neutralhead
Arm & Tokens & C share & S share & U share & L share & Mass \\
\midrule
Ordinary & 0.3B & 25.88 & 25.35 & 24.68 & 24.09 & 194.42 \\
Ordinary & 1.2B & 9.80 & 49.63 & 2.64 & 37.92 & 103.40 \\
Ordinary & 3B & 4.67 & 51.95 & 1.42 & 41.96 & 82.27 \\
Ordinary & 100B & 2.56 & 44.84 & 0.00 & 52.60 & 60.00 \\
\addlinespace[2.5pt]
Curriculum & 0.3B & 25.80 & 25.00 & 25.67 & 23.53 & 199.02 \\
Curriculum & 1.2B & 2.96 & 54.92 & 2.17 & 39.95 & 90.56 \\
Curriculum & 3B & 4.63 & 53.08 & 3.79 & 38.50 & 86.82 \\
Curriculum & 100B & 0.42 & 47.93 & 0.03 & 51.61 & 53.93 \\
\bottomrule\end{tabular}
\end{center}
\end{table}

The severity-cutoff sweep preserves the same ordering: Circuit contributes substantially early on but little at the endpoint, where Store and Local account for most remaining mass (Table~\ref{tab:audit-sensitivity}).

\begin{table}[!htbp]
\caption{Selected 942-target conflict cells. Arms: O = Ordinary, C = Curriculum. The C/S/U columns report Circuit/Store/Use shares of actionable excess-NLL mass; Mass is unnormalized.}
\label{tab:residual-selected}
\begin{center}
\papertable
\setlength{\tabcolsep}{4pt}
\begin{tabular}{lrrrrr@{\hspace{12pt}}lrrrrr}
\toprule
\neutralhead
Arm & Tokens & C & S & U & Mass & Arm & Tokens & C & S & U & Mass \\
\midrule
O & 0.100 & 35.74 & 11.34 & 52.91 & 2430.94 & C & 0.100 & 33.53 & 11.87 & 54.60 & 2277.94 \\
O & 0.481 & 12.27 & 10.60 & 77.12 & 2132.58 & C & 0.481 & 7.78 & 8.14 & 84.08 & 2058.97 \\
O & 0.800 & 5.75 & 9.39 & 84.86 & 1420.16 & C & 0.800 & 3.40 & 10.67 & 85.93 & 1209.27 \\
O & 1.200 & 4.74 & 15.19 & 80.08 & 1136.97 & C & 1.200 & 1.25 & 17.21 & 81.53 & 926.17 \\
O & 3.000 & 1.99 & 23.23 & 74.77 & 740.32 & C & 3.000 & 5.74 & 23.93 & 70.33 & 771.53 \\
O & 30.00 & 2.93 & 13.39 & 83.68 & 536.51 & C & 30.00 & 2.35 & 13.85 & 83.80 & 585.74 \\
O & 100.0 & 2.86 & 15.77 & 81.37 & 500.96 & C & 100.0 & 1.44 & 12.02 & 86.54 & 532.32 \\
\bottomrule
\end{tabular}
\end{center}
\end{table}

The conflict panel contains 560 H--H and 382 H--L targets and is Use-heavy by construction. Its 100B percentages are conditional residual composition, while the separately frozen natural roster measures case transitions below.

\begin{table}[!htbp]
\caption{Adjudication-free first-checkpoint audit on 272 fixed identities. Mass is absolute excess NLL above 2; solved means full-view target rank $=1$.}
\label{tab:prospective-transition}
\begin{center}
\papertable
\begin{tabular}{llrrrr}
\toprule
\neutralhead
Arm & Tokens & Circuit cases & Circuit mass & Circuit share & Total mass / solved \\
\midrule
Ordinary & 0.3B & 37 & 227.10 & 13.7\% & 1658.47 / 0 \\
Curriculum & 0.3B & 50 & 318.19 & 19.0\% & 1673.17 / 0 \\
Ordinary & 0.6B & 9 & 41.80 & 3.1\% & 1364.17 / 7 \\
Curriculum & 0.6B & 1 & 8.14 & 0.7\% & 1243.58 / 20 \\
Ordinary & 1.2B & 0 & 0.00 & 0.0\% & 906.40 / 66 \\
Curriculum & 1.2B & 0 & 0.00 & 0.0\% & 834.39 / 94 \\
Ordinary & 100B & 2 & 5.74 & 1.0\% & 596.87 / 103 \\
Curriculum & 100B & 0 & 0.00 & 0.0\% & 589.97 / 109 \\
\bottomrule
\end{tabular}
\end{center}
\end{table}

For the checkpoint-local audit in Figure~\ref{fig:diagnosis}a, 272 identities are frozen from actionable Ordinary@0.3B behavior and replayed under both arms. No adjudication labels enter the subsequent classification. The enclosing 360-case roster was retrospectively enriched using cross-checkpoint strata, so the audit establishes within-roster dynamics, not natural-token prevalence. Circuit contracts under either Use-first or Local-first precedence; most early Circuit cases are solved by 100B (Tables~\ref{tab:prospective-transition} and~\ref{tab:audit-sensitivity}). These are changes in residual demand, not a requirement that each fact visit all three labels.

\begin{table}[!htbp]
\caption{Residual-audit sensitivity. The threshold-rescaling audit, natural-case severity sweep, and adjudication-free trajectory use their own frozen populations. Percentage shares refer to excess-loss mass.}
\label{tab:audit-sensitivity}
\begin{center}\papertable
\setlength{\tabcolsep}{4pt}
\begin{tabular}{@{}P{0.40\textwidth}P{0.56\textwidth}@{}}
\toprule\neutralhead
Audit or quantity & Setting or observed range \\
\midrule
Positive strong-rule threshold multipliers & $0.5,\ 0.75,\ 1.25,\ 1.5$ \\
Changed assignments & 0.8--3.1\% \\
Actionable Circuit / Use shares & 18.0--19.2\% / 80.8--82.0\% \\
\midrule
Natural-case severity cutoffs & $0,\ 1,\ 2,\ 3,\ 4$ \\
Ordinary Circuit share at 0.3B & 25.7--26.3\% across the cutoff grid \\
Ordinary Circuit share at 100B & 7.1, 4.6, 2.6, 0, 0\%, in cutoff order \\
Endpoint Store + Local share & 89--100\% \\
\midrule
Adjudication-free audit & Thresholds 0--0.5; Use-first / Local-first \\
Early Circuit mass clears by & 1.2B under both precedence orders \\
Early Circuit cases solved at 100B & Ordinary: 31/37; Curriculum: 43/50 \\
\bottomrule
\end{tabular}
\end{center}
\end{table}

\subsection{Same-fact Web-to-QA counterfactual}

The paired panel contains 435 formally reviewed Web failures: 289 semantic questions and 146 cloze items. Each retains its fact, answer span, and target subtoken across presentation views. No-context analysis uses the 241 semantic questions independently judged to identify the same fact without the passage; passage-dependent references and cloze slots are excluded from that view. Aggregate values are reweighted to the frozen Web residual strata used to sample the panel.

\begin{table}[!htbp]
\caption{Same-fact presentation counterfactual. NLL is target NLL; gap is Curriculum-minus-Ordinary. Top-1 reports Ordinary/Curriculum accuracy (\%).}
\label{tab:web-qa}
\begin{center}\papertable
\setlength{\tabcolsep}{5pt}
\begin{tabular}{lrrrrr}
\toprule\neutralhead
View & Cases & Ordinary NLL & Curriculum NLL & Gap & Target top-1 \\
\midrule
Original Web continuation & 435 & 4.821 & 5.719 & 0.898 & 31.3 / 16.6 \\
Context QA & 435 & 5.288 & 5.575 & 0.288 & 45.1 / 42.1 \\
Paraphrased context QA & 435 & 5.115 & 5.444 & 0.329 & 46.9 / 43.4 \\
Minimal-source QA & 435 & 4.556 & 4.877 & 0.321 & 54.0 / 52.4 \\
Source-masked QA & 435 & 7.300 & 7.587 & 0.287 & 19.3 / 15.6 \\
Shuffled-context QA & 435 & 7.893 & 8.041 & 0.148 & 11.7 / 8.5 \\
No-context QA & 241 & 7.304 & 7.567 & 0.263 & 10.8 / 7.5 \\
\bottomrule\end{tabular}\end{center}
\end{table}

The Curriculum--Ordinary gap narrows substantially when the same facts are presented as context QA (Table~\ref{tab:web-qa}). Relevant evidence helps both arms relative to shuffled context, and masking the correct source removes much of that benefit. Both models therefore use the explicit source; natural Web continuation adds demands of surface realization, source identification, and output selection. Factual/entity and long-tail phrase cases show the largest context-QA reduction, whereas pointer, morphology, and local-relation cases retain more of the original gap.

\clearpage
\section{Circuit Evidence}
\label{app:circuit}

\begin{figure}[!htbp]
  \centering
  \includegraphics[width=\textwidth]{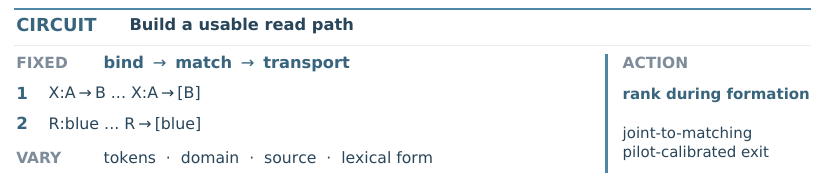}
  \caption{\textbf{Circuit data construction.} Preserve key--match--successor roles while varying content and source realization. Formation-time sensitivity selects natural rows; joint binding--matching followed by matching-only selection orders prerequisites; a pilot calibrates the exit.}
  \label{fig:circuit-construction}
\end{figure}

\subsection{Functional construct}

For head $h$, let $\alpha_h(i,j)$ be attention from position $i$ to $j$, $q$ the query, $v^+$ and $v^-$ the correct and decoy value positions, and $(k_r,v_r)_{r=1}^2$ the source pairs. The binding offset $d=v_r-k_r$ is shared by both pairs within a prompt and varies across prompts; $\mathcal D$ is the candidate offset set. The clean probe reports binding selectivity $A_h$, matching selectivity $B_h$, and value transport $C_h$:
\begin{align}
A_h &= \frac12\sum_{r=1}^{2}\alpha_h(v_r,k_r)
      - \frac{1}{|\mathcal D\setminus\{d\}|}
        \sum_{\delta\in\mathcal D\setminus\{d\}}
        \frac12\sum_{r=1}^{2}\alpha_h(v_r,v_r-\delta),\\
B_h &= \alpha_h(q,v^+)-\alpha_h(q,v^-),\\
C_h &= \left(W_h^O V_h(v^+)-W_h^O V_h(v^-)\right)^\top
       \left(u_{\mathrm{tgt}}-u_{\mathrm{dec}}\right).
\end{align}
Here $u_{\mathrm{tgt}}$ and $u_{\mathrm{dec}}$ are the unembedding vectors of the correct and decoy tokens. Development and test each contain 128 clean--corrupt pairs of length 192; offset, source distance, slot, query, and token family vary independently. Development selects heads and freezes thresholds; test reports clean--corrupt and component-specific causal gates.

\begin{table}[!htbp]
\caption{Auxiliary 160M functional confirmation. Top: domain-probe binding, matching, and transport controls, where $\Delta$NLL is ablated minus clean. Bottom: natural-carrier confirmation at 100M training tokens in the main and independent runs. Every registered gate passes.}
\label{tab:abc-gates}
\label{tab:abc-100m-confirm}
\begin{center}
\papertable
\setlength{\tabcolsep}{2.5pt}
\begin{tabular}{@{}P{0.085\linewidth}N{0.135\linewidth}N{0.08\linewidth}N{0.08\linewidth}N{0.105\linewidth}N{0.14\linewidth}N{0.095\linewidth}N{\dimexpr0.24\linewidth-35pt\relax}@{}}
\toprule
\circuithead
Domain & \multicolumn{2}{c}{Binding} & \multicolumn{3}{c}{Matching} & \multicolumn{2}{c}{Transport} \\
\cmidrule(lr){2-3}\cmidrule(lr){4-6}\cmidrule(lr){7-8}
& Clean--corrupt & Patch & Clean & Swapped & Q/K $\Delta$NLL & Eligible & V/O $\Delta$NLL \\
\midrule
Web  & .00703 & .00575 & .00958 & $-.11722$ & .09406 & 86  & .35515 \\
Math & .00748 & .00326 & .04658 & $-.15676$ & .72399 & 139 & .68819 \\
Code & .00690 & .00222 & .06866 & $-.15794$ & 1.18290 & 124 & 1.14056 \\
\bottomrule
\end{tabular}
\par\vspace{5pt}
\setlength{\tabcolsep}{2.5pt}
\begin{tabular}{@{}P{0.235\linewidth}P{0.125\linewidth}N{0.105\linewidth}N{0.105\linewidth}N{0.105\linewidth}N{0.105\linewidth}N{\dimexpr0.18\linewidth-30pt\relax}@{}}
\toprule
\circuithead
Run & Domain & \multicolumn{3}{c}{Matching} & Binding & Transport \\
\cmidrule(lr){3-5}\cmidrule(lr){6-6}\cmidrule(lr){7-7}
& & Clean & Accuracy & Swapped & Ablation drop & Gain \\
\midrule
Main 100M & Web  & .042 & .688 & $-.163$ & .045 & 1.398 \\
Main 100M & Math & .069 & .828 & $-.169$ & .069 & 1.763 \\
Main 100M & Code & .098 & .836 & $-.253$ & .100 & 2.725 \\
\addlinespace[2.5pt]
Independent 100M & Web  & .041 & .758 & $-.117$ & .042 & 1.109 \\
Independent 100M & Math & .048 & .852 & $-.134$ & .044 & 1.333 \\
Independent 100M & Code & .093 & .766 & $-.329$ & .093 & 2.087 \\
\bottomrule
\end{tabular}
\end{center}
\end{table}

The top panel uses a separate domain-probe pool with 256 clean/corrupt/swapped cases per domain, each of length 512. Binding here is a residual key-readout margin; Patch is its patched-minus-corrupt recovery after inserting clean head activations. Eligible counts cases, not heads: clean binding margin exceeds $-0.03191$, matching accuracy exceeds $0.5$, and matching selectivity exceeds $0.0006303$. These are the frozen development gates for this pool, distinct from the 128-pair formal construct above.

Both auxiliary runs reproduce positive clean matching selectivity, negative swap reversal, positive binding-ablation drop, and positive transport gain in every domain (Table~\ref{tab:abc-100m-confirm}). The target-scale trajectory runs a fresh dynamic scan at each checkpoint: top-ranked head identities change during formation, with no overlap between the early scan and the endpoint top set. A fixed-reference audit preserves the early Curriculum matching advantage but not the endpoint advantage. Table~\ref{tab:circuit-settings} records the scan and reference checkpoints. Together, these results track formation of a functional path across scales, not one immutable head set.

\subsection{Teacher and organization}

The formation-time and endpoint-time MDA arms select equal-sized, partially overlapping pools (Table~\ref{tab:circuit-settings}). Target initialization, optimizer, slots, budget, selected component, and matching-only organization are fixed. MDA uses empirical-Fisher EK-FAC~\citep{george2018ekfac} over the prescribed attention Q/K blocks; scores are converted to within-domain percentile ranks before selection.

\begin{table}[!htbp]
\caption{Strict matching-only comparison of formation-time and endpoint-time MDA teachers. Lower is better.}
\label{tab:mda-teacher}
\begin{center}
\papertable
\begin{tabular}{r rr rr}
\toprule
\circuithead
& \multicolumn{2}{c}{Overall NLL} & \multicolumn{2}{c}{Ambiguous recall NLL} \\
\cmidrule(lr){2-3}\cmidrule(lr){4-5}
Tokens & Formation & Endpoint & Formation & Endpoint \\
\midrule
0.1B & 7.58435 & \textbf{7.57876} & 6.48897 & \textbf{6.46127} \\
0.2B & \textbf{6.28988} & 6.30441 & \textbf{4.97248} & 4.97939 \\
0.3B & \textbf{5.70661} & 5.74712 & \textbf{4.37319} & 4.42689 \\
0.45B & \textbf{5.10654} & 5.11048 & 3.71070 & \textbf{3.66996} \\
0.6B & \textbf{4.58137} & 4.59874 & 2.75410 & \textbf{2.74828} \\
0.9B & \textbf{4.05616} & 4.11359 & \textbf{2.00536} & 2.06120 \\
1.2B & \textbf{3.79557} & 3.82161 & \textbf{1.72919} & 1.74228 \\
\bottomrule
\end{tabular}
\end{center}
\end{table}

Formation-time MDA lowers NLL in every domain at the comparison endpoint, confirming that the formation checkpoint is the stronger teacher for this matching-targeted intervention (Figure~\ref{fig:circuit}c).

\begin{table}[!htbp]
\caption{Normalized AUC over 0.1--1.2B for the static teacher-by-organization factorial. Teacher denotes the matching-score checkpoint; the binding-score checkpoint is fixed.}
\label{tab:mda-factorial}
\begin{center}
\papertable
\setlength{\tabcolsep}{2.5pt}
\begin{tabular}{@{}P{0.11\linewidth}P{0.20\linewidth}N{0.12\linewidth}N{0.12\linewidth}N{0.11\linewidth}N{0.12\linewidth}N{\dimexpr0.22\linewidth-30pt\relax}@{}}
\toprule
\circuithead
Teacher & Selection order & Binding $\uparrow$ & Matching $\uparrow$ & Induction $\uparrow$ & \shortstack[r]{Overall\\NLL $\downarrow$} & \shortstack[r]{Ambiguous\\recall NLL $\downarrow$} \\
\midrule
Formation & \jointb{} & .7128 & \textbf{.3071} & \textbf{2.4977} & 4.8303 & \textbf{3.0933} \\
Formation & Matching-only & .7084 & .2964 & 2.4410 & \textbf{4.8223} & 3.0960 \\
Endpoint & \jointb{} & .6996 & .2903 & 2.4098 & 4.8392 & 3.1095 \\
Endpoint & Matching-only & \textbf{.7141} & .2709 & 2.2298 & 4.8513 & 3.1117 \\
\bottomrule
\end{tabular}
\end{center}
\end{table}

Formation-time MDA improves the relevant AUCs within both organizations, and \jointb{} improves matching, induction, and ambiguous recall within both teachers (Table~\ref{tab:mda-factorial}). An independent update-utility audit also favors the formation teacher before the path forms, then changes sign afterward. Its sample and checkpoint boundary are recorded in Table~\ref{tab:circuit-settings}.

\subsection{Order, exit, and persistence}

\begin{table}[!htbp]
\caption{Same selected warm-up multiset, different order, evaluated at 2.5B. All columns report PPL except ambiguous recall NLL; lower is better.}
\label{tab:order-control}
\begin{center}
\papertable
\setlength{\tabcolsep}{4pt}
\begin{tabular}{@{}P{0.25\linewidth}N{0.13\linewidth}N{0.13\linewidth}N{0.12\linewidth}N{0.17\linewidth}N{\dimexpr0.20\linewidth-40pt\relax}@{}}
\toprule
\circuithead
Order & Overall & Math & Code & Ambiguous recall NLL & Web \\
\midrule
\rowcolor{circuitwash}\jointb{} & \textbf{7.6686} & \textbf{18.3830} & \textbf{4.0426} & \textbf{0.6923} & 41.4245 \\
Interleaved & 7.6872 & 18.4269 & 4.0561 & 0.6955 & 41.3716 \\
Matching-to-joint & 7.7018 & 18.4238 & 4.0703 & 0.7026 & \textbf{41.2755} \\
\bottomrule
\end{tabular}
\end{center}
\end{table}

\begin{table}[!htbp]
\caption{Warm-up exit sweep at the common 3B endpoint. Values are PPL; lower is better.}
\label{tab:exit-sweep}
\begin{center}
\papertable
\setlength{\tabcolsep}{4.5pt}
\begin{tabular}{lrrrrrrr}
\toprule
\circuithead
Exit & Overall & Code & STEM & Books & Papers & English Web & Chinese Web \\
\midrule
Ordinary & 25.9247 & 3.7900 & 17.2590 & 29.6611 & 21.5048 & 35.3020 & 70.5480 \\
0.3B & 25.8479 & 3.7790 & 17.2387 & 29.4727 & 21.5418 & 35.2636 & 70.0989 \\
0.6B & 25.7401 & 3.7798 & 17.2283 & 29.2783 & 21.5368 & \textbf{35.1533} & 69.3255 \\
\rowcolor{circuitwash}1.2B & \textbf{25.6834} & \textbf{3.7703} & \textbf{17.1703} & \textbf{29.2310} & 21.4604 & 35.1711 & \textbf{68.9548} \\
2.0B & 25.7771 & 3.7780 & 17.2160 & 29.4766 & \textbf{21.4597} & 35.2412 & 69.2234 \\
\bottomrule
\end{tabular}
\end{center}
\end{table}

\begin{table}[!htbp]
\caption{Selected validation PPL from the paired 100B run.}
\label{tab:dense-100b}
\begin{center}
\papertable
\begin{tabular}{rrr}
\toprule
\circuithead
Tokens & Ordinary & Curriculum \\
\midrule
75B & 12.6560 & \textbf{12.6019} \\
80B & 12.6177 & \textbf{12.5621} \\
82B & 12.6043 & \textbf{12.5497} \\
90B & 12.5714 & \textbf{12.5164} \\
95B & 12.5625 & \textbf{12.5078} \\
100B & 12.5580 & \textbf{12.5034} \\
\bottomrule
\end{tabular}
\end{center}
\end{table}

Every evaluation in the dense late-training window favors Curriculum, as does every endpoint domain (Figure~\ref{fig:circuit}d; Tables~\ref{tab:dense-100b} and~\ref{tab:100b-domains}). The earlier crossing of Ordinary's final PPL is therefore followed by a sustained advantage. Together with the formation trajectory, teacher comparison, fixed-multiset order test, exit sweep, and independent-initialization comparison, this persistence completes the Circuit evidence chain.

\begin{table}[!htbp]
\caption{Curriculum versus Ordinary across independent 3B initializations. Gains use $100(P_{\rm Ord}-P_{\rm Curr})/P_{\rm Ord}$.}
\label{tab:3b-replication}
\begin{center}\papertable
\begin{tabular}{lrl}
\toprule\circuithead
Run & Overall PPL gain (\%) & Ambiguous recall NLL direction \\
\midrule
1 & 1.26 & lower \\
2 & 1.58 & lower by 0.00740 \\
3 & 1.00 & higher by 0.00321 \\
\bottomrule\end{tabular}
\end{center}
\end{table}

\begin{table}[!htbp]
\caption{All six domain PPLs at the matched 100B endpoint.}
\label{tab:100b-domains}
\begin{center}\papertable
\begin{tabular}{lrr}
\toprule\circuithead
Domain & Ordinary & Curriculum \\
\midrule
Books & 15.4227 & \textbf{15.3798} \\
Chinese Web & 21.6994 & \textbf{21.3875} \\
Code & 2.4922 & \textbf{2.4882} \\
English Web & 16.9887 & \textbf{16.9355} \\
Papers & 11.2392 & \textbf{11.2190} \\
STEM & 9.2482 & \textbf{9.2225} \\
\bottomrule\end{tabular}\end{center}
\end{table}

\section{Store Evidence}
\label{app:store}

\begin{figure}[!htbp]
  \centering
  \includegraphics[width=\textwidth]{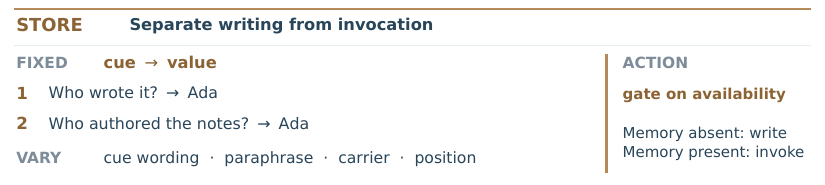}
  \caption{\textbf{Store data construction.} Preserve target content and the cue--value relation; vary cue surface, carrier, and optimizer position. Evidence rescue after M-only failure calls for writing; M-only success followed by carrier suppression calls for invocation training.}
  \label{fig:store-construction}
\end{figure}

\subsection{Gate and timescale}

\begin{paperalgorithm}{M-missing gate and operation-matched action}{alg:memory-action}
\algline{1}{\textbf{Input:} no-source/M-only, evidence-present, and local/template-control views.}
\algline{2}{Select M-missing when the no-source candidate decision fails, the evidence-present decision succeeds, and evidence rescue passes the fixed threshold.}
\algline{3}{For M-missing, preserve the cue--value relation and vary cues, paraphrases, natural carriers, and optimizer steps.}
\algline{4}{If M-only succeeds but both natural carriers suppress its margin, route the case to gated \muse{} rather than repeated writing.}
\algline{5}{Audit exact relation, strict reverse, unseen paraphrase, wrong pair, non-drag controls, and broad-language cost.}
\end{paperalgorithm}

The Store roster is frozen without using future gain as a selection criterion. Its target NLL continues to improve substantially more than that of refined Circuit or Use cases (Table~\ref{tab:store-contrasts}). The set overlaps rare n-grams, fixed forms, morphology, entities, facts, and local relations, so prolonged learning is a heterogeneous coverage result rather than a single memory type.

\begin{table}[!htbp]
\caption{Availability-gated and generic hard-example populations, each with 2,048 selected cases. The gated set is frozen from the 2,220 qualifying M-missing candidates. Evidence-present success is part of that selection rule.}
\label{tab:m-missing-gate}
\begin{center}
\papertable
\begin{tabular}{lrr}
\toprule
\storehead
Selector & Frozen cases & Evidence-present accuracy \\
\midrule
H/L/M M-causal gate & 2,048 & \textbf{100.00\%} \\
Blind/hard & 2,048 & 31.84\% \\
Random hard & 2,048 & 5.08\% \\
\bottomrule
\end{tabular}
\end{center}
\end{table}

The gate filters natural QA candidates, including TriviaQA \citep{joshi2017triviaqa}, by no-source failure and evidence rescue; Table~\ref{tab:intervention-settings} records the retained populations. This operation gate concentrates content-resolvable failures more effectively than generic hard-example selection (Table~\ref{tab:m-missing-gate}).

\subsection{Views and selection}
\label{app:store-views}

The natural QA audit compares an original answer with a type-matched replacement. A valid item has distinct answers, a source containing the original answer but not the replacement, and a usable first divergent answer token. At that position, candidate success means $m_\theta>0$; target-span NLL instead averages all answer tokens. The shared answer prefix is supplied when candidates diverge after the first token.

\begin{table}[!htbp]
\caption{Store input views. The question and candidate answers stay fixed. Evidence availability and answer-free context change separately.}
\label{tab:store-views}
\begin{center}\papertable
\begin{tabular}{@{}P{0.19\textwidth}P{0.76\textwidth}@{}}
\toprule\storehead
View & Input construction \\
\midrule
M-only & Short factual question followed by \texttt{Answer:}; no source passage. \\
Evidence & Original answer-containing passage followed by the same question. \\
Surface & The source sentence truncated immediately before its answer; scored as a completion. \\
Carrier A / B & The same question preceded by background with answer-containing sentences removed; A uses a Background instruction, B a Notes instruction. \\
\bottomrule
\end{tabular}
\end{center}
\end{table}

The reference M-missing selector first removes M-only successes, then requires evidence-present success and a margin rescue of at least one logit unit. It ranks eligible items by rescue for fixed-size selection. The surface view diagnoses sentence-prefix effects separately. Blind selection ranks no-source failures by their negative margin; Random hard greedily matches the selected margins using failures outside the M-missing pool.

The reference invocation selector starts from M-only successes. It requires mean drag at least $0.5$ and drag at least $0.1$ under each carrier, then ranks by mean drag. Controls are M-available facts with mean drag at most $0.1$. Greedy matching selects the closest M-only margin among remaining controls of the same entity type and answer-token length, falling back to the remaining pool when that stratum is empty. Controls are used without replacement. Table~\ref{tab:intervention-settings} records the frozen evaluation populations.

\subsection{Writing and natural use}

The controlled novel-relation panel contains 128 distinct assignments and uses tokenizer-native keys, multiple cues, strict reversed pairs, unseen paraphrases, and wrong-pair training. Packed cue--value segments use block-diagonal attention and independent position IDs; only target spans contribute loss, the target does not appear in the cue, and cue views recur across separate optimizer steps. The total continuation is fixed at 24 optimizer steps; $k$ use \mwrite{} and the remainder restore the same ordinary stream.

\begin{table}[!htbp]
\caption{Controlled relation-writing dose. Margin gains are relative to the parent; broad NLL cost is relative to ordinary continuation.}
\label{tab:m-write-dose}
\begin{center}
\papertable
\begin{tabular}{rrrr}
\toprule
\storehead
Write steps $k$ & Exact margin & Unseen paraphrase & Broad NLL cost \\
\midrule
12 & +0.01296 & +0.02679 & +0.227\% \\
14 & +0.03729 & +0.02959 & +0.298\% \\
15 & +0.05009 & +0.02910 & +0.328\% \\
16 & +0.05744 & +0.03181 & +0.358\% \\
20 & +0.07162 & +0.03394 & +0.674\% \\
\bottomrule
\end{tabular}
\end{center}
\end{table}

The strict-reverse margin exchanges the same two candidates and is the algebraic negative of the exact margin. Unseen paraphrases and the separately trained wrong-pair arm provide the relation controls: correct pairing improves both cue views, whereas wrong pairing favors the swapped assignment (Table~\ref{tab:relation-controls}). The panel thus tests assignment learning on fixed keys and transfer to a held-out cue realization. These controls identify a directional preference for the intended assignment across cue forms. At this dose, the margin response is clearer than candidate-choice accuracy, which does not improve consistently across the two views; the reported result is relation learning measured by margin.

\begin{table}[!htbp]
\caption{Relation controls at the low-dose operating point ($k=14$). Entries are no-source correct-minus-competing margins; wrong-pair training changes the main-relation margin by $-0.0536$ while favoring the swapped assignment.}
\label{tab:relation-controls}
\begin{center}\papertable
\begin{tabular}{lrrr}
\toprule\storehead
Cue view & Parent & Correct pairing & Wrong pairing \\
\midrule
Exact & $-0.07528$ & $-0.03799$ & $-0.12884$ \\
Unseen paraphrase & $-0.03448$ & $-0.00490$ & $-0.04868$ \\
\bottomrule
\end{tabular}
\end{center}
\end{table}

For Equation~\ref{eq:memory-diagnosis}, $m_\theta(u;x)=\log p_\theta(y^+\mid x)-\log p_\theta(y^-\mid x)$ compares the correct and competing tokens at their first divergent answer position. Natural-carrier margin is the mean over two fixed carriers. Drag change is continuation-minus-parent margin drag; negative values mean less suppression. The two gain columns instead use parent-minus-continuation target-span NLL, averaged over the same facts (and both carriers for natural use).

For natural use, we select a frozen set from M-available cases with strict circuit drag and cross dragged-versus-control write and use slots. Ordinary rows, tokens, optimizer positions, and continuation budget remain fixed. Table~\ref{tab:intervention-settings} gives the population and slot configuration.

\begin{table}[!htbp]
\caption{Natural Write$\times$Use factorial. Gains are parent-minus-continuation NLL; larger is better. A/B mean is the natural-use gain in Table~\ref{tab:store-main}. Use trains carrier A; B changes the prompt around the same answer-free background and frozen facts. All arms share two intervention slots and a 25M-token continuation, with matched non-drag facts in disabled slots. Parent, ordinary rows, loss, optimizer positions, updates, and budget are fixed.}
\label{tab:m-write-use}
\begin{center}
\papertable
\setlength{\tabcolsep}{4pt}
\begin{tabular}{llrrrrr}
\toprule
\storehead
Arm & Active operation & M-only & Carrier A & Carrier B & A/B mean & Drag change \\
\midrule
C0 & matched controls & 0.636 & 0.901 & 0.887 & 0.894 & $-0.927$ \\
C1 & Write & 0.740 & 0.975 & 0.967 & 0.971 & $-0.741$ \\
C2 & Use & 0.723 & \textbf{1.213} & \textbf{1.189} & \textbf{1.201} & \textbf{$-1.814$} \\
C3 & Write + Use & \textbf{0.806} & \textbf{1.307} & \textbf{1.282} & \textbf{1.294} & $-1.751$ \\
\bottomrule
\end{tabular}
\end{center}
\end{table}

Use contributes the larger gain under each carrier, including the alternative prompt B, while the interaction remains small (Table~\ref{tab:m-write-use}). Matched non-drag controls show why the diagnosis matters: an invocation intervention that helps suppressed facts need not help facts without that failure. Table~\ref{tab:store-contrasts} records these contrasts separately from the dragged-fact factorial.

\begin{table}[!htbp]
\caption{Store audit summaries for separate populations. The first block reports natural-use NLL effects on dragged facts; the second reports drag change on matched non-drag controls (negative means less suppression). The final block reports target-NLL decreases from 3B to 100B on frozen class rosters.}
\label{tab:store-contrasts}
\begin{center}\papertable
\begin{tabular}{llr}
\toprule\storehead
Population & Contrast & Value \\
\midrule
Dragged & Write main effect & 0.085 \\
Dragged & Use main effect & 0.315 \\
Dragged & Write$\times$Use interaction & 0.016 \\
\midrule
Non-drag control & Write-only drag change & $-0.358$ \\
Non-drag control & Use-only drag change & $+0.393$ \\
\midrule
Frozen Store (Store41) & Target-NLL decrease & 1.903 \\
Refined Circuit & Target-NLL decrease & 0.297 \\
Refined Use & Target-NLL decrease & 0.225 \\
Local & Target-NLL decrease & 0.712 \\
\bottomrule
\end{tabular}
\end{center}
\end{table}

\paperneedspace{22\baselineskip}
\section{Use Evidence}
\label{app:use}

\begin{figure}[!htbp]
  \centering
  \includegraphics[width=\textwidth]{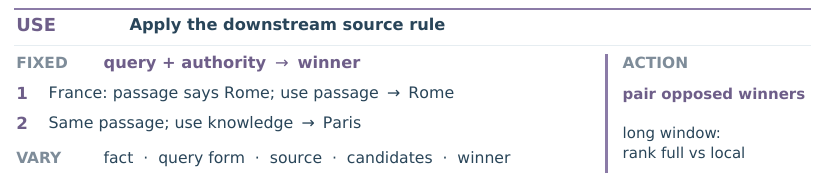}
  \caption{\textbf{Use data construction.} In the illustrative conflict, the passage says Rome while established knowledge says Paris. Changing task authority changes the prescribed answer. Preserve that conditional relation across facts and query forms; long-window selection instead concentrates full--local opportunity. Table~\ref{tab:paired-construction} gives paired, unpaired, and reversed constructions.}
  \label{fig:use-construction}
\end{figure}

\subsection{Relation engagement and transfer}

\paragraph{A worked construction.}
Table~\ref{tab:paired-construction} shows the base three-template construction with shortened illustrative facts. Each fact record specifies one question, an edited passage, and two competing answers. A deterministic family hash chooses template $t$ for true pairing; the unpaired arm uses the other two templates and alternates its assigned authority mode across records. Template $T_0$ asks either to ``Use the passage as the authority'' or to answer ``from established knowledge rather than copying the substitution.'' $T_1$ uses Text/Query/Response fields, and $T_2$ uses Excerpt/Answer fields, with corresponding passage-authorized and knowledge-authorized instructions.

\begin{table}[!htbp]
\caption{Illustrative paired, unpaired, and reversed training records. $C$ authorizes the passage and $M$ established knowledge. Each listed arrow gives the supervised target; the other answer is its competitor. For readability, both illustrative records are assigned $t=T_0$. The construction follows the implemented allocation; the facts illustrate it rather than report model outputs.}
\label{tab:paired-construction}
\begin{center}\papertable
\setlength{\tabcolsep}{4pt}
\begin{tabular}{@{}P{0.08\textwidth}P{0.42\textwidth}P{0.21\textwidth}P{0.23\textwidth}@{}}
\toprule\usehead
Fact & Question & Edited passage & Knowledge \\
\midrule
A & What is the capital of France? & Capital: Rome & Paris \\
B & What is the capital of Germany? & Capital: Vienna & Berlin \\
\bottomrule
\end{tabular}
\vspace{5pt}
\begin{tabular}{@{}P{0.20\textwidth}P{0.37\textwidth}P{0.37\textwidth}@{}}
\toprule\usehead
Training arm & Fact A: two decisions & Fact B: two decisions \\
\midrule
True paired & $T_0,C\to$ Rome; $T_0,M\to$ Paris & $T_0,C\to$ Vienna; $T_0,M\to$ Berlin \\
Balanced unpaired & $T_1,C\to$ Rome; $T_2,C\to$ Rome & $T_1,M\to$ Berlin; $T_2,M\to$ Berlin \\
Reversed & $T_0,C\to$ Paris; $T_0,M\to$ Rome & $T_0,C\to$ Berlin; $T_0,M\to$ Vienna \\
\bottomrule
\end{tabular}
\end{center}
\end{table}

Both true and unpaired stores contain two decisions per record and equal total counts of the two authority modes; they differ in within-fact mode coverage and template allocation. Ordinary training rows and supplementary loss weights are shared. Familiar evaluation applies all three training templates to unseen families; independent evaluation uses two separately written templates. Training excludes held-out family identities and their original/replacement entities. Thus the familiar-query comparison tests learning the conditional relation across facts, while the independent-query panel tests transfer of that relation across wording.

The positive-control panel fixes 1,000 unseen natural fact families while retaining familiar training query surfaces. True pairing, balanced unpaired carriers, Ordinary, and a reversed relation share the continuation budget and carrier structure.

\begin{table}[!htbp]
\caption{Familiar-query positive control over 1,000 unseen fact families. Accuracy is reported as a percentage; strict family requires every registered surface for a family.}
\label{tab:paired-engagement}
\begin{center}
\papertable
\begin{tabular}{llrrrr}
\toprule
\usehead
Checkpoint & Arm & Override & Reject & Strict pair & Strict family \\
\midrule
+100M & Ordinary & 73.57 & 23.23 & 4.27 & 0.10 \\
+100M & Balanced unpaired & 73.47 & 34.70 & 9.43 & 1.00 \\
+100M & True paired & 73.53 & 39.67 & \textbf{13.83} & \textbf{1.90} \\
\addlinespace[2.5pt]
+150M & Ordinary & 74.77 & 22.63 & 4.47 & 0.00 \\
+150M & Balanced unpaired & 74.10 & 35.37 & 10.60 & 0.80 \\
+150M & True paired & 77.13 & 37.03 & \textbf{14.87} & \textbf{2.90} \\
\bottomrule
\end{tabular}
\end{center}
\end{table}

On an independent matched relation panel, true pairing strengthens the trained route offset and reversing the relation changes its sign (Table~\ref{tab:use-diagnostics}). This directional response complements the balanced-exposure comparison.

\begin{table}[!htbp]
\caption{Strict-pair accuracy (\%) under independently phrased queries. Each repair is evaluated at two saved points.}
\label{tab:paired-failures}
\begin{center}
\papertable
\begin{tabular}{lrrrr}
\toprule
\usehead
Round & Ordinary@100M & Paired@100M & Ordinary@150M & Paired@150M \\
\midrule
3 templates & 7.00 & 5.45 & 6.55 & 4.55 \\
32 templates & 2.35 & \textbf{2.95} & 2.60 & \textbf{2.70} \\
10$\times$ families & 6.70 & 2.95 & 6.60 & 2.95 \\
Pair-worst & 6.70 & 3.75 & 6.60 & 3.85 \\
\bottomrule
\end{tabular}
\end{center}
\end{table}

The independent surfaces show that familiar-query pairing is insufficient for broader transfer (Table~\ref{tab:paired-failures}). They identify query realization and opposed winner direction as dimensions to target in the next conditional-arbitration construction.

\subsection{Context opportunity ranking}

The \ccur{} score is defined in Equation~\ref{eq:ccur}. We score successive target blocks using either the full window or their preceding local context; Table~\ref{tab:long-context-settings} gives the window and block sizes. The two truncated views retain identical tokens and either reset positions to zero ($L$) or retain their original offsets ($R$). The score averages the clipped minimum loss response over all targets, including zeros. These coordinate conventions implement the same context cut under standard relative-position RoPE; the minimum is the executed scoring rule. LongCE and \ccur{} match source part, domain, split, full-window-NLL strata, objective, update count, and token budget, isolating the association between full--local response and selected window identity.

Within each source shard, the high-full-NLL half of training rows forms the candidate pool. We stratify by difficulty, rank candidates by \ccur{}, and select distinct documents under a fixed per-shard quota. LongCE matches each anchor by full-window NLL while excluding that exact window; almost all pairs occupy the same difficulty bin. On the original pool, \ccur{} beats LongCE at every registered aggregate length and checkpoint. The independent replication excludes every original training document and gives all arms equal exposure to the new pool. Table~\ref{tab:long-context-settings} records the counts and matching configuration.

\begin{table}[!htbp]
\caption{Document-disjoint data-pool replication. All arms share the parent, LongCE objective, matched strata, 92 updates, and 96.47M tokens. Lower answer-token NLL is better.}
\label{tab:ccur-independent}
\begin{center}
\papertable
\begin{tabular}{llrrr}
\toprule
\usehead
Checkpoint & Arm & 4K & 8K & 12K \\
\midrule
Midpoint & LongCE & 4.2271 & 4.5152 & 4.4247 \\
Midpoint & Score-shuffled & 4.1894 & 4.4917 & 4.4060 \\
Midpoint & \ccur{} & \textbf{4.1479} & \textbf{4.4368} & \textbf{4.3580} \\
\midrule
Endpoint & LongCE & 3.5081 & 3.7562 & 3.8365 \\
Endpoint & Score-shuffled & 3.4721 & 3.7069 & 3.7906 \\
Endpoint & \ccur{} & \textbf{3.4103} & \textbf{3.6264} & \textbf{3.7179} \\
\bottomrule
\end{tabular}
\end{center}
\end{table}

The score-shuffled control preserves the shard/bin score multiset, shard totals, unique-document cap, and budget while breaking the score-to-window relation. \ccur{} beats both controls at every length and checkpoint (Table~\ref{tab:ccur-independent}); paired task-stratified 95\% endpoint intervals exclude zero throughout (Table~\ref{tab:ccur-ci}). Overall endpoint PPL also favors \ccur{} (Table~\ref{tab:use-diagnostics}).

Evaluation uses a fixed synthetic RULER-style \citep{hsieh2024ruler} validation set, balanced across word--number retrieval in noise (\texttt{niah\_single\_1}), retrieval among needles (\texttt{niah\_multikey\_2}), UUID retrieval (\texttt{niah\_multikey\_3}), and four-hop variable tracking (\texttt{vt}). We teacher-force the gold answers and microaverage NLL over answer tokens. Case and token counts appear in Table~\ref{tab:long-context-settings}. Document disjointness concerns the training pools; all arms share the frozen synthetic evaluation cases.

\begin{table}[!htbp]
\caption{Paired endpoint uncertainty for the document-disjoint training pool. Positive control-minus-\ccur{} NLL favors \ccur{}. Intervals resample complete evaluation cases within task, preserving answer-token aggregation (50,000 draws).}
\label{tab:ccur-ci}
\begin{center}\papertable
\begin{tabular}{llrr}
\toprule\usehead
Length & Control & NLL gain & 95\% interval \\
\midrule
4K & LongCE & 0.0979 & [0.0928, 0.1031] \\
4K & Score-shuffled & 0.0618 & [0.0574, 0.0663] \\
8K & LongCE & 0.1298 & [0.1223, 0.1374] \\
8K & Score-shuffled & 0.0806 & [0.0741, 0.0872] \\
12K & LongCE & 0.1186 & [0.1109, 0.1264] \\
12K & Score-shuffled & 0.0727 & [0.0665, 0.0790] \\
\bottomrule\end{tabular}\end{center}
\end{table}

\subsection{Conditional-arbitration audit}

\begin{table}[!htbp]
\caption{Use-localization controls on the original pool. Values are \ccur-minus-LongCE truth-log-probability gains.}
\label{tab:ccur-localization}
\begin{center}
\papertable
\begin{tabular}{lrrrrrr}
\toprule
\usehead
View/test & Full & Local only & Full$-$local & Source removal & Swap winner & Swap value \\
\midrule
Gain & .00789 & $-.00634$ & \textbf{.01424} & $-.00080$ & $-.00490$ & $-.00761$ \\
\bottomrule
\end{tabular}
\end{center}
\end{table}

Complete context localizes the average first-pool gain, but source removal and value swaps do not support a semantic winner rule (Table~\ref{tab:ccur-localization}). Target-level localization also does not replicate on the document-disjoint models; the repeated positive result is answer likelihood and candidate ranking.

\begin{table}[!htbp]
\caption{Fixed 1,000-family context-versus-memory panel. Accuracy is reported as a percentage; Both requires both query modes for one family.}
\label{tab:hm-boundary}
\begin{center}
\papertable
\begin{tabular}{llrrr}
\toprule
\usehead
Checkpoint & Arm & Context override & Memory reject & Both \\
\midrule
Midpoint & LongCE & 38.55 & 59.60 & 1.70 \\
Midpoint & \ccur{} & 38.90 & 58.65 & 1.15 \\
\addlinespace[2.5pt]
Endpoint & LongCE & 40.55 & 55.85 & 1.30 \\
Endpoint & \ccur{} & 41.10 & 55.00 & 1.20 \\
\bottomrule
\end{tabular}
\end{center}
\end{table}

At endpoint, improved context override comes at the expense of memory-authorized decisions; no registered comparison improves both modes and pair completion (Table~\ref{tab:hm-boundary}). The domain-matched BridgeSet likewise shows broad gains without Use-specific improvement (Table~\ref{tab:use-diagnostics}). These controls delimit the positive result as task-aligned window ranking and a source-prior shift, not general H/L/M arbitration.

\begin{table}[!htbp]
\caption{Additional Use diagnostics. Blocks report separate populations and metrics: an independent matched relation panel at +100M, Overall PPL at the document-disjoint endpoint, and a domain-matched 96-case BridgeSet. BridgeSet entries are NLL gains; specificity subtracts the mean gain of the other classes.}
\label{tab:use-diagnostics}
\begin{center}\papertable
\begin{tabular}{llr}
\toprule\usehead
Audit & Arm or class & Value \\
\midrule
Route offset & Ordinary & 0.0282 \\
 & True paired & 0.1009 \\
 & Reversed relation & $-0.0884$ \\
\midrule
Overall PPL & LongCE & 19.3240 \\
 & \ccur{} & 19.1569 \\
\midrule
BridgeSet NLL gain & Circuit & .00281 \\
 & Store & .02539 \\
 & Use & .01471 \\
 & Local & .01729 \\
 & Use specificity & $-.00046$ \\
\bottomrule
\end{tabular}
\end{center}
\end{table}

\paperneedspace{10\baselineskip}
\section{Reproducibility}
\label{app:reproducibility}

All quantitative figures are generated by \texttt{scripts/plot\_figures.py} from checked CSV/JSON aggregates under \texttt{figures/source\_data}. The script emits vector PDFs and inspection PNGs and asserts sample counts, arm/checkpoint identities, mass conservation, and plotted comparison directions. Figure~\ref{fig:overview}, the three construction schematics, and the sequential branch diagram explain task structure or protocol; no image model creates or reconstructs quantitative marks.

The materialized candidate pool is shared across the primary arms (Table~\ref{tab:circuit-settings}). Text is normalized, filtered by frozen benchmark substrings, exact-deduplicated, and conservatively near-deduplicated by deterministic LSH bands. The source package includes LaTeX/BibTeX, style files, plotting code, aggregate figure inputs, and vector/PNG figures. It does not distribute corpus text, checkpoints, or every raw per-example array; source licensing, privacy, and redistribution review remain separate from this scientific reproducibility record.

\paperneedspace{9\baselineskip}
\section{Detailed Intervention Protocols}
\label{app:method-details}

\paragraph{Conflict-classifier precedence.}
The classifier first checks validity, availability, and whether a failure is actionable. The nonactionable label means the causal-actionability gate does not fire, rather than requiring a rank-one answer. Strong counterfactual rules then take precedence over frozen adjudication; an ordered fallback handles the remaining cases (Table~\ref{tab:classifier-rules}). Recognized adjudications map selection, local competition, use/readout, binding, and parametric coverage to their corresponding fine classes; artifact and unresolved-mechanism judgments map to ambiguous/invalid. Circuit groups binding, Store groups parametric coverage, and Use groups selection, arbitration, and value-use. In this conflict panel, Store enters through recognized parametric-coverage adjudication, not an automated M-only fallback; unadjudicated rows cannot receive Store from these rules. The separate natural-memory gate directly evaluates M-only availability. Only actionable classes contribute excess mass. Missing source-removal effects default to zero.

\begin{table}[!htbp]
\caption{Frozen conflict-classifier rules. Quantities $a,c,w,l$ are defined in Appendix~\ref{app:protocol}; $r_F$ and $r_L$ are full and best-local target ranks, and $m_F$ is the full-view margin against the strongest wrong candidate. Within each ordered block, apply the first matching rule. Frozen adjudication uses 360 representative records.}
\label{tab:classifier-rules}
\begin{center}\papertable
\setlength{\tabcolsep}{3 pt}
\begin{tabular}{@{}P{0.07\textwidth}P{0.67\textwidth}P{0.21\textwidth}@{}}
\toprule\neutralhead
Stage & Condition & Result \\
\midrule
Gate & Empty target, replacement character, or no alphanumeric character & Invalid \\
 & $ a\geq 0.25$ or source-only target rank $=1$ & Available source \\
 & $ r_F\geq 2$ or full NLL $\geq 2$ or $ m_F<0$ & Bad output \\
 & Bad output and $[w\geq 0.05$ or $ c\geq 0.05$ or $(l\geq 0.10\land r_L<r_F)]$ & Actionable \\
 & Valid but nonactionable & Nonactionable \\
\midrule
\multicolumn{3}{l}{\textbf{Strong rules, in precedence order}} \\
1 & High-confidence frozen artifact judgment & Invalid \\
2 & H--H; available source; $ m_F<0$; $ w\geq 0.5$ & Selection \\
3 & $ m_F<0$; $ w\geq 0.2$; $ l\geq 0.5$ & Local--global arbitration \\
4 & Available source; $ m_F\geq 0$; $ r_F\geq 2$; $ a\geq 1$ & Value-use \\
5 & H--H; unavailable source; $\max(a,c)<0.25$ & Availability/binding \\
6 & $ l\geq 1$; $ w<0.2$ & Local--global arbitration \\
7 & Recognized high/medium-confidence frozen adjudication & Mapped fine class \\
\midrule
\multicolumn{3}{l}{\textbf{Fallback rules, in precedence order}} \\
1 & Available source; $ m_F<0$; $ w\geq 0.05$ & Selection \\
2 & $l\geq 0.10$ and $r_L<r_F$ & Arbitration \\
3 & Unavailable source; bad output & Binding \\
4 & Available source; $ m_F<0$; $ c<0.05$ & Ambiguous \\
5 & Available source; $ m_F\geq 0$; $ r_F\geq 2$ & Value-use \\
6 & Otherwise & Ambiguous \\
\bottomrule
\end{tabular}
\end{center}
\end{table}

\paragraph{Teacher and prefix construction.}
Binding and matching scores use checkpoints appropriate to their formation times; the endpoint teacher supplies the mature comparison. The strict matching-only test holds Q/K components fixed and refits EK-FAC factors at each teacher checkpoint (Table~\ref{tab:circuit-settings}). Within each domain, let $ a_z,b_z$ be binding and matching score percentiles. Joint binding--matching selection ranks by $\min(a_z,b_z)$ and matching-only selection by $ b_z$. The prefix places joint selection first, then concentrates matching demand, while preserving domain slots and using distinct natural sequences.

\paragraph{Exit calibration and the long run.}
The target-scale pilot varies the curriculum exit and follows each prefix with ordinary data to a shared endpoint (Table~\ref{tab:exit-sweep}). Formation requires consistent binding, matching, and read behavior at adjacent checkpoints, positive ambiguous recall loss under dynamically located head ablation, persistence into ordinary continuation, and held-out Overall/Web performance. The best validation exit occurs after the earliest passing formation checkpoint. The long run fixes that schedule in advance, then follows ordinary training through 100B tokens. The pilot therefore supplies a measured formation window, while the long run tests a frozen schedule derived from it; exact operating points are in Table~\ref{tab:circuit-settings}.

\begin{table}[!htbp]
\caption{Circuit data and protocol settings. These are fixed experimental instances of the construction described in the text; teacher checkpoints and target-model checkpoints are distinct.}
\label{tab:circuit-settings}
\begin{center}\papertable
\setlength{\tabcolsep}{4 pt}
\begin{tabular}{@{}P{0.30\textwidth}P{0.66\textwidth}@{}}
\toprule\circuithead
Item & Setting \\
\midrule
Teacher parameters & 349,746,176 \\
Teacher checkpoints & Binding: 200.278M tokens; matching formation: 1.501B; endpoint: 3.000B \\
Matching components & Teacher Q/K heads L9H5 and L9H13 \\
Prefix allocation & Joint selection 40\%; matching-only 60\% \\
Primary prefix & 234,701 joint + 352,051 matching-selected sequences; 1.2017B tokens \\
Domain slots & Code 10\%, STEM 10\%, Books 25\%, Papers 15\%, English Web 25\%, Chinese Web 15\% \\
Validation & 37,718,637 shared target positions \\
Materialized candidate pool & 60,951,155 packed sequences; 124,827,965,440 target tokens \\
Teacher-time selected pools & 585,728 sequences/arm; 280,560 shared; Jaccard 0.3149 \\
\midrule
Dynamic head audit & No 0.3B head remains in the endpoint top-eight set \\
Fixed-reference audit & Ordinary@0.8B; early advantage through 1.2B, absent at 100B \\
Matching update-utility audit & 384 natural sequences; formation-minus-endpoint utility positive through 0.6B, then changes sign \\
Pilot exits / endpoint & 0.3, 0.6, 1.2, 2B / 3B tokens \\
Persistence requirement & At least 0.5B ordinary tokens after the candidate exit \\
Earliest passing exit & 0.6B \\
Frozen long-run prefix & 1.2B; best pilot Overall PPL at 3B \\
Dense late-training audit & 16 evaluations from 80B through 100B \\
\bottomrule
\end{tabular}
\end{center}
\end{table}

\paragraph{Memory treatment and control slots.}
The natural factorial pairs dragged facts with matched non-drag controls. Each arm repeats fixed Write and Use slots at the same optimizer positions (Table~\ref{tab:intervention-settings}). Active slots draw dragged facts; disabled slots draw matched non-drag facts in the same format. Write presents the cue--value relation, while Use presents its natural carrier. Independently attended segments each carry one supervised decision target. The arm definitions in Table~\ref{tab:m-write-use} specify which operation is active; remaining updates use the same ordinary rows. All arms resume the same model and optimizer. Training and evaluation share the frozen facts. Write uses the sentence prefix ending at the answer; Use trains carrier A, while evaluation reports M-only and both A/B prompts around the same answer-free background. The intervention loss marks only the first divergent answer token, whereas the reported NLL gains average the target span. The controlled novel-relation study separately holds out a cue realization while retaining the learned assignments.

\paragraph{Paired route objective and evaluation units.}
A training pair contains context-authorized and memory-authorized decisions for one fact family (Table~\ref{tab:paired-construction}). Balanced unpaired supervision uses the same fact records and two decisions per record, assigns only one mode to each record, and balances modes across records. It uses the other two query templates; true pairing uses both modes of one template. Reversed supervision keeps the true-paired inputs and exchanges the target and competitor labels at the decision positions. Ordinary next-token loss is supplemented by $\lambda[\mathrm{softplus}(z^- - z^+)-\log p_\theta(y^+\mid x)]$, where $z^+,z^-$ are target and competitor logits at the first divergent answer token and $p_\theta$ is the full-vocabulary softmax. The supplementary loss is additive over decisions. Table~\ref{tab:intervention-settings} gives $\lambda$ and the evaluation units. A decision is correct when the authorized candidate wins at its first divergent token. Strict pair requires both authority modes for a row; strict family requires all registered surfaces. With authorized-minus-competing margins $m_C$ and $m_M$ for context and memory modes, route offset is $\mathbb E[(m_C+m_M)/2]$. It measures signed relation alignment, while strict-pair accuracy requires both decisions to succeed. Independent phrasings change the query surface, while the additional controls vary template diversity, family coverage, and emphasis on the less successful direction (Table~\ref{tab:paired-failures}). The final saved checkpoint follows an ordinary-training retention stage.

\begin{table}[!htbp]
\caption{Store and paired-Use protocol settings. Gate candidates, natural invocation cases, and route-evaluation families are separate populations.}
\label{tab:intervention-settings}
\begin{center}\papertable
\setlength{\tabcolsep}{4 pt}
\begin{tabular}{@{}P{0.31\textwidth}P{0.65\textwidth}@{}}
\toprule\storehead
Item & Setting \\
\midrule
M-missing gate & 12,000 candidates; 11,997 valid; 2,220 M-missing \\
Natural-use filtering & 11,468 complete-view candidates; 7,247 M-available; 2,212 strict-drag cases \\
Natural factorial & 512 dragged facts + 512 matched non-drag controls \\
Slot schedule & Write and Use slots, each repeated 3 times among 24 updates \\
Packed row & Four independently attended 512-token segments; one supervised target/segment \\
Parent / learning rate & Same 100B model and optimizer; $2\times 10^{-5}$ \\
\midrule\usehead
\multicolumn{2}{l}{Paired-route training and evaluation} \\
\midrule
Base relation store & 8,192 rows; 16,384 decision targets; 8,192 per authority mode \\
Relation schedule & Every second update during the 96-update relation stage \\
Objective coefficient & $\lambda=0.00075$ for each supplementary term \\
Familiar-query evaluation & 1,000 unseen families; 3 surfaces/family; 3,000 rows \\
Independent-query evaluation & 2 surfaces/family; 2,000 rows \\
Retention checkpoint & +100M relation stage, then 50M ordinary tokens; final +150M \\
\bottomrule
\end{tabular}
\end{center}
\end{table}

\paragraph{Long-context objective and extension.}
All long-context arms extend the same 2K parent to 16K and resume under matched optimization (Table~\ref{tab:long-context-settings}). LongCE~\citep{fang2025longppl} uses full-context next-token loss with detached weights $ w_i=\min\{\exp(\ell_i^L-\ell_i^F),5\}$ on scored positions and unit weights on the unscored local-history prefix. The loss is the mean of $ w_i\ell_i^F$ over target tokens. The local teacher uses the selector's same history/block construction. The training weight uses the reset-position local view $L$; the original-position view $R$ supplies the additional guard in the selection score, not a second training weight. LongCE, \ccur{}, and score-shuffled arms share this objective and context extension; selection changes their complete training windows.

\begin{table}[!htbp]
\caption{Long-context scoring, selection, and evaluation settings. The independent pool excludes all original training documents. All arms within a pool share the stated configuration.}
\label{tab:long-context-settings}
\begin{center}\papertable
\setlength{\tabcolsep}{4 pt}
\begin{tabular}{@{}P{0.31\textwidth}P{0.65\textwidth}@{}}
\toprule\usehead
Item & Setting \\
\midrule
Full window / scored positions & 16,384 tokens / 4,096--16,383 ($12,288$ targets) \\
Target blocks / local history & Twelve 1,024-target blocks / 4,096 preceding tokens \\
Context extension & 2K$\rightarrow$16K; linear RoPE scaling by 8~\citep{chen2023positioninterpolation} \\
Learning rate / batch & $2\times 10^{-5}$ / 64 complete windows per update \\
\midrule
Original window pool & 126,170 windows across 8 source shards \\
Difficulty strata & High-full-NLL half of each shard; 20 equal-frequency bins \\
Original selection & 768 distinct-document windows/shard; 6,144 total \\
LongCE anchor matches & 6,105 same-bin pairs; 39 adjacent-bin pairs \\
Original training budget & 100.66M tokens per arm \\
Independent pool exclusion & All 10,542 original documents \\
Independent selection & 184 windows/shard; 1,472 total; 4 equal exposures \\
Independent continuation & 92 updates; 96.47M tokens \\
\midrule
Evaluation lengths & 4K, 8K, 12K \\
Evaluation cases & 800/length; 200/task \\
Answer tokens by length & 10,597 / 10,594 / 10,668 \\
\bottomrule
\end{tabular}
\end{center}
\end{table}

\clearpage
\section{Sequential Validation}
\label{app:sequential}

\subsection{One continuous training chain}

The sequential study uses an approximately 350M-parameter architecture and one tokenizer. Routes A (ordinary order) and B (early Circuit curriculum) start from the same initialization. Their pretraining data multiset, per-position domain allocation, optimizer, and learning-rate schedule are matched. They are parallel routes: B does not inherit the teacher's model parameters. B uses 458 Joint updates, 686 Matching updates, and 1,717 ordinary updates; A uses ordinary order throughout. This prefix schedule is fixed before the common-task endpoints are evaluated.

\begin{figure}[!htbp]
\centering
\includegraphics[width=\textwidth]{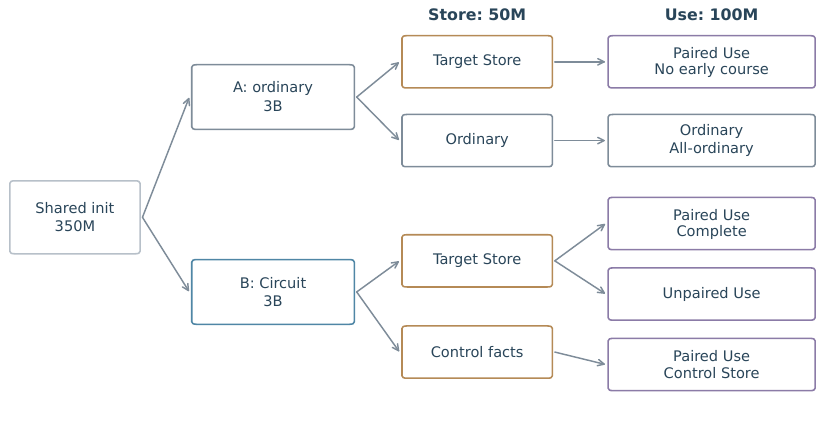}
\caption{\textbf{Continuous branches with a shared initialization.} A/B differ in early pretraining order. Target and control Store write matched but different facts. The paired/unpaired comparison shares a target-Store parent. Only arrows denote model-state inheritance.}
\label{fig:sequential-flow}
\end{figure}

Store inherits its route's model, optimizer, and random state. Use inherits the corresponding Store state; branches with the same parent start from the same saved state. Figure~\ref{fig:sequential-flow} shows the four Store parents and five final routes. Fixed schedules test composition of the interventions, rather than online switching between them.

\begin{table}
\caption{Sequential training settings. Positions are model-input positions, not parameter counts or GPU costs.}
\label{tab:sequential-settings}
\begin{center}\papertable
\begin{tabular}{lrrrr}
\toprule\neutralhead
Stage & Updates & Effective batch & Input positions & Learning rate \\
\midrule
Pretraining (each A/B) & 2,861 & 512 & 2,999,975,936 & shared schedule \\
Store (each parent) & 381 & 64 & 49,938,432 & $3\times10^{-5}$ \\
Use (ordinary stream) & 96 & 512 & 100,663,296 & $2\times10^{-5}$ \\
\bottomrule
\end{tabular}
\end{center}
\end{table}

All ordinary sequences have 2,048 positions. Store keeps stage budget, ordinary-text schedule, teaching slots, and exposures per fact fixed. Use keeps ordinary inputs, auxiliary-input count, teaching facts, and effective updates fixed. The ordinary-continuation control receives the same auxiliary forward inputs with zero relation weight; this matches input volume, not auxiliary-backward compute. The two trunks, four Store parents, and five Use endpoints total approximately 6.71B input positions including auxiliary inputs.

\subsection{Facts, supervision, and evaluation}

The target pool contains 1,280 facts selected using A's pretraining endpoint: the model fails the no-evidence candidate decision and succeeds when correct evidence is supplied. Another 1,280 real facts provide matched controls by answer type, text length, and format. Target membership is not reselected using B or continuation outcomes. The comparison is therefore on this A-selected common pool.

\begin{table}
\caption{Fixed target-fact split. The main sequential results use only the 640 common tracking facts.}
\label{tab:sequential-split}
\begin{center}\papertable
\begin{tabular}{lrrr}
\toprule\neutralhead
Partition & Facts & Store-taught & Use-taught \\
\midrule
Use teaching & 384 & yes & yes \\
Common tracking & 640 & yes & no \\
Development & 256 & yes & no \\
\bottomrule
\end{tabular}
\end{center}
\end{table}

Target Store writes the target questions and answers; control Store uses matched control questions and answers. Each fact receives 19 teaching exposures at distinct updates. A sequence contains four independently attended QA segments, with supervision at the first token where the correct and competing answers differ. Teaching and ordinary updates follow a fixed interleaving schedule.

For Use, each teaching fact has a true answer and a passage with a conflicting answer. One instruction authorizes the passage; the other authorizes established knowledge. Paired and balanced-unpaired constructions each contain 384 facts, 384 relation rows, and 768 decisions, balanced across the two authority modes. Paired teaching gives both modes for one fact and template; unpaired teaching gives one mode per fact using the other two base templates. Thus fact count, decision count, overall mode balance, and loss are matched, while within-fact authority coverage and template allocation differ.

Use adds target cross-entropy and target-versus-competitor ranking losses, each weighted by $0.00075$, to ordinary language modeling. Every second ordinary update includes eight relation rows (16 decisions), for 48 auxiliary batches. Ordinary and auxiliary gradients enter the same optimizer update, with accumulation preserving the effective batch.

Every node is evaluated on identical common facts and queries: one correct-evidence reading question per fact, four no-evidence memory wordings per fact, and three familiar authority templates with two decisions each. The latter gives 1,920 pairs and 3,840 decisions. Repeated templates are not additional independent facts. Answer NLL sums loss over all tokens of the full correct answer and divides by the total answer-token count in that view. Joint-use NLL pools both authority modes, so a lower average does not imply improvement in each mode. Pair accuracy requires the authorized candidate to win at its first divergent token under both instructions; it is not complete-answer generation accuracy.

\subsection{Stage and endpoint results}

\begin{table}
\caption{All 18 stage-trajectory values on the same 640 common facts. A uses ordinary pretraining; B uses the early Circuit curriculum. Both then receive target Store and paired Use. Lower answer-token NLL is better.}
\label{tab:sequential-stages}
\begin{center}\papertable
\begin{tabular}{llrrr}
\toprule\neutralhead
Route & Node & Reading NLL & Memory NLL & Joint-use NLL \\
\midrule
A & Pretraining & 3.156392 & 6.025372 & 3.863549 \\
A & Store & 2.179583 & 4.426465 & 3.192157 \\
A & Use & 2.170958 & 4.371545 & 3.050034 \\
\midrule
B & Pretraining & 3.136392 & 5.995372 & 3.839549 \\
B & Store & 2.139583 & 4.346465 & 3.132157 \\
B & Use & 2.130958 & 4.291545 & 2.970034 \\
\bottomrule
\end{tabular}
\end{center}
\end{table}

\begin{table}
\caption{Within-stage NLL reductions on the common facts. The final column is the difference of the two observed improvements, not a fitted interaction or significance estimate.}
\label{tab:sequential-deltas}
\begin{center}\papertable
\begin{tabular}{lrrr}
\toprule\neutralhead
Contrast & A & B & B minus A \\
\midrule
Store: memory before minus after & 1.599 & 1.649 & 0.050 \\
Use: joint-use before minus after & 0.142 & 0.162 & 0.020 \\
\bottomrule
\end{tabular}
\end{center}
\end{table}

\begin{table}
\caption{All five final routes on the common task. Pair-NLL win rate sums the two full-answer NLLs within a query pair and counts when the complete route is lower than the listed control; ties are not wins. Its denominator is 1,920 pairs, and it is distinct from candidate accuracy.}
\label{tab:sequential-full}
\begin{center}\papertable
\setlength{\tabcolsep}{4pt}
\begin{tabular}{lrrrr}
\toprule\neutralhead
Route & Memory NLL & Joint-use NLL & Pair accuracy & Pair-NLL wins \\
\midrule
Complete sequence & 4.292 & \textbf{2.970} & \textbf{5.26\%} & --- \\
No early curriculum & 4.372 & 3.050 & 3.39\% & 58.28\% \\
Control facts in Store & 5.220 & 3.245 & 1.41\% & 79.74\% \\
Unpaired Use & \textbf{4.289} & 3.019 & 3.18\% & 55.10\% \\
Ordinary continuation & 6.026 & 3.870 & 0.99\% & 91.77\% \\
\bottomrule
\end{tabular}
\end{center}
\end{table}

The complete sequence improves the common use task over every replacement, including no early curriculum. Unpaired teaching has slightly lower memory NLL but weaker joint use, separating answer availability from task-authorized selection. The absolute pair accuracy remains low; the positive result is the sequential improvement under familiar instructions, consistent with the broader arbitration frontier in Section~\ref{sec:use}.

\subsection{A common-fact handoff}

The worked fact asks when Newton died. Established knowledge gives 1727; the fixed conflict passage reads: ``Newton died in 1790. Poems honoring his scientific work continued to appear for decades after his death.'' Passage-authorized queries require 1790; knowledge-authorized queries require 1727. This fact is Store-taught and Use-untaught. Table~\ref{tab:sequential-case} follows the same fixed question and candidate pair at each node. It is an explanatory case, not an estimate of how often this sequence of errors occurs.

\begin{table}
\caption{One fact across the continuous chain and controls. Margins are authorized-minus-competing first-divergent-token scores in the reading and no-evidence views. Pair NLL sums the two authority decisions for the fixed query pair.}
\label{tab:sequential-case}
\begin{center}\papertable
\setlength{\tabcolsep}{4pt}
\begin{tabular}{lrrrr}
\toprule\neutralhead
Node or control & Reading margin & Memory margin & Pair NLL & Both correct \\
\midrule
A pretraining & +1.375 & $-0.469$ & 5.144 & no \\
B pretraining & +1.450 & $-0.375$ & 5.100 & no \\
B + target Store & +6.250 & +8.938 & 3.300 & no \\
Complete sequence & +5.062 & +7.031 & 2.850 & yes \\
No early curriculum & +4.188 & +5.312 & 3.011 & no \\
Unpaired Use & +4.312 & +6.469 & 2.975 & no \\
\bottomrule
\end{tabular}
\end{center}
\end{table}

Reading already succeeds before Store, but unaided recall does not. Store establishes the correct memory candidate without resolving the conflict pair; subsequent paired Use resolves that pair. This illustrates the distinction between a usable read path, available content, and query-conditioned control without requiring every fact to follow an identical trajectory.

\end{document}